\documentclass[lettersize,journal]{IEEEtran}
\usepackage{times}

\usepackage[numbers]{natbib}
\usepackage{multicol}
\usepackage[bookmarks=true]{hyperref}
\usepackage[]{xcolor}
\usepackage[]{bm}
\usepackage{amsmath,amsfonts,amsthm,amsbsy,amssymb}
\usepackage{graphicx}
\usepackage{float}
\usepackage{placeins}
\usepackage[font=small,labelfont=bf]{caption}
\usepackage{booktabs}
\usepackage{tabularx}
\usepackage{tabularray}
\usepackage{titlesec}
\usepackage{etoolbox}
\usepackage{physics}
\usepackage{changepage}

\newcommand{\rev}[1]{{\color{black}#1}}
\def\nvar{n}
\def\ncon{m}
\def\nparam{d}
\def\bx{\bm{x}}
\def\blam{\bm{\lambda}}
\def\bz{\bm{z}}

\newcommand{\vect}[1]{{\rm vec}(#1)}
\newcommand{\inner}[2]{\left<#1,#2\right>}

\newcommand{\indset}[1]{\left[#1\right]}
\newcommand{\bth}{\bm{\theta}}

\def\solnmap{\mathcal{S}}
\def\feasset{\bm{\Omega}_{\bth}}

\DeclareMathOperator{\img}{im}
\DeclareMathOperator{\corank}{corank}
\DeclareMathOperator{\codim}{codim}

\newtheorem{theorem}{Theorem}
\newtheorem{lemma}[theorem]{Lemma}
\newtheorem{definition}[theorem]{Definition}
\newtheorem{proposition}[theorem]{Proposition}
\newtheorem{remark}{Remark}

\newcounter{ProblemCounter}
\newenvironment{problem}[1]{%
  \refstepcounter{ProblemCounter}%
  \begin{equation}%
  \label{#1}%
}{%
  \tag{P\theProblemCounter}%
  \end{equation}%
}

\def\footnoterev#1{\footnote{\rev{#1}}}

\begin{document}

\title{SDPRLayers: Certifiable Backpropagation Through Polynomial Optimization Problems in Robotics}

\author{Connor Holmes, Frederike D{\"u}mbgen, Timothy D. Barfoot\vspace*{-0.45in}
\thanks{Connor Holmes and Timothy D. Barfoot are with the University of Toronto Robotics Institute, University of Toronto, Toronto, Ontario, Canada, \texttt{connor.holmes@mail.utoronto.ca}, \texttt{tim.barfoot@utoronto.ca}; Frederike D{\"u}mbgen is with Inria, École Normale Supérieure, PSL University, Paris, France, \texttt{frederike.duembgen@gmail.com}.}}%




\setlength\intextsep{0pt}
\newcommand{\pretablespace}{5pt}
\newenvironment{response}
    {
	\begin{adjustwidth}{3em}{0em}\itshape
    }
    { 
    \end{adjustwidth}
    }

\maketitle

\begin{abstract}
	\rev{
	A recent set of techniques in the robotics community, known as \emph{certifiably correct methods}, frames robotics problems as \emph{polynomial optimization problems} (POPs) and applies convex, semidefinite programming (SDP) relaxations to either find or certify their global optima. 
	In parallel, \emph{differentiable optimization} allows optimization problems to be embedded into end-to-end learning frameworks and has received considerable attention in the robotics community. In this paper, we consider the ill effect of convergence to spurious local minima in the context of learning frameworks that use differentiable optimization. We present SDPRLayers, an approach that seeks to address this issue by combining convex relaxations with implicit differentiation techniques to provide \emph{certifiably correct solutions and gradients} throughout the training process. We provide theoretical results that outline conditions for the correctness of these gradients and provide efficient means for their computation. Our approach is first applied} to two simple-but-demonstrative simulated examples, which expose the potential pitfalls of \rev{reliance on local optimization in} existing, state-of-the-art, differentiable optimization methods. We then apply our method in a real-world application: we train a deep neural network to detect image keypoints for robot localization in challenging lighting conditions. We provide our open-source, PyTorch implementation of SDPRLayers\footnote{Code available at \url{https://github.com/utiasASRL/sdprlayer}} and our differentiable localization pipeline\footnote{Code available at \url{https://github.com/utiasASRL/deep_learned_visual_features/tree/mat-weight-sdp-version}}.
\end{abstract}



\section{Introduction}\label{sec:introduction}

\IEEEPARstart{T}{he} versatility of learning models has made them ubiquitous in robotics, \rev{permeating} almost all layers of the modern software stack~\cite{wangPyPoseLibraryRobot2023}. On the other hand, model-based optimization -- the mainstay of traditional robotics -- provides a level of robustness, accuracy, and generalization that has proven difficult to match by learning-based methods~\cite{romeroActorCriticModelPredictive2023}. 
A \rev{recent} paradigm, leveraging advances in so-called \textit{differentiable optimizers}, now enables roboticists to combine these two approaches into a single \emph{end-to-end learning} framework.

In this approach, optimization problems are embedded as `layers' in deep-learning networks, with optimization parameter data as the input and the optimal solution as the output. Similar to other layers in machine learning, the forward pass of the layer involves solving the optimization, while the backward pass computes the gradients. 

The benefits of this approach are several, allowing practitioners to capitalize on the respective advantages of model-based and learning approaches while also mitigating their disadvantages. For example, this integration means that domain knowledge can be injected directly into the robotics pipeline, while still allowing machine learning parameters to be trained on the final goal of a robotics module. This also obviates costly integration and fine-tuning stages that are necessary when learning and optimization modules are developed in parallel~\cite{wangPyPoseLibraryRobot2023}.

\begin{figure}[t]
    \centering
    \includegraphics[width=\linewidth]{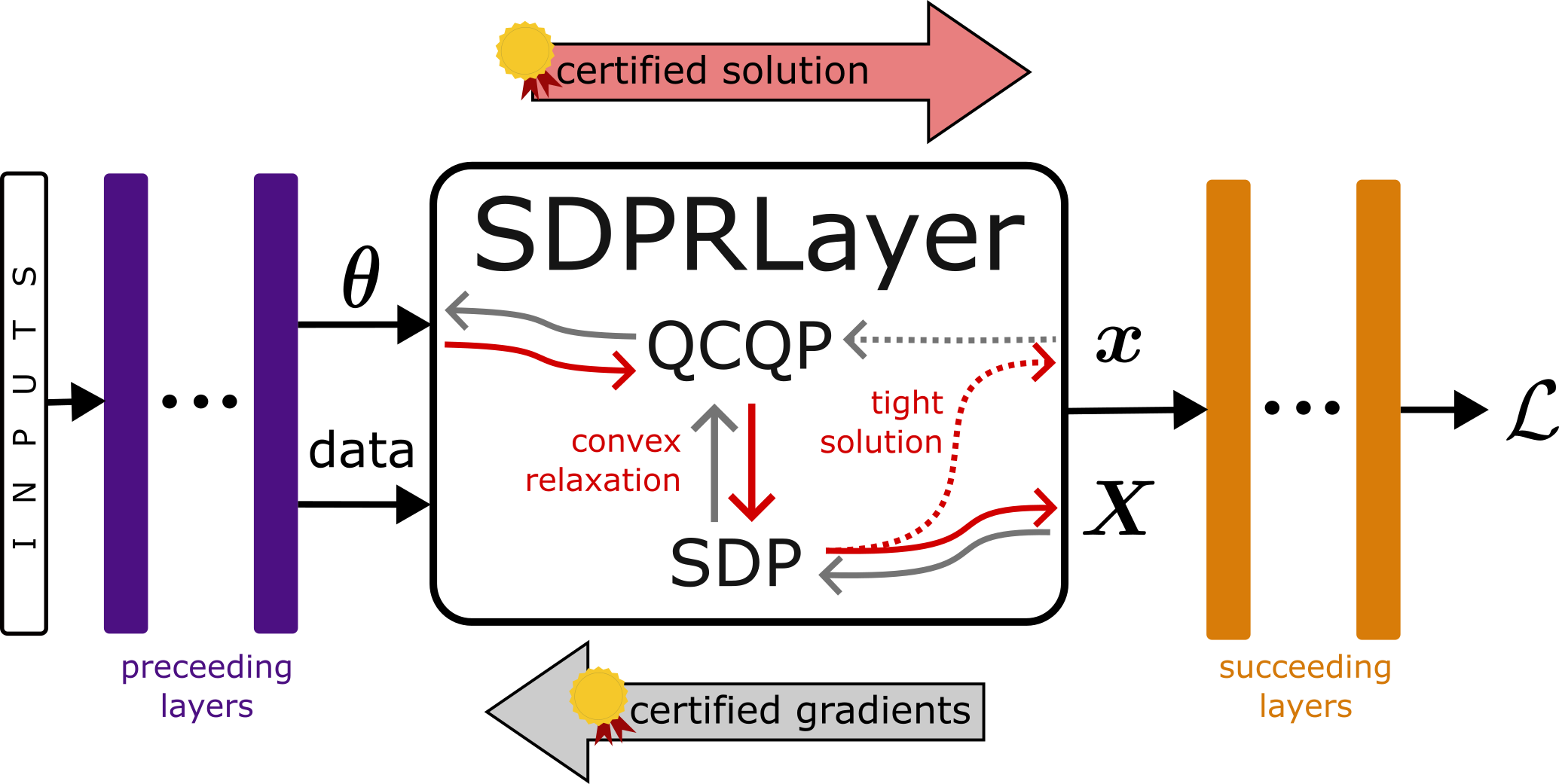}
    \caption{Our SDPRLayer embedded in a PyTorch autodifferentiation graph. For non-convex problems with \emph{tight} semidefinite relaxations, the SDPRLayer finds the certified, globally optimal solution \rev{and makes it differentiable via implicit differentiation of the non-convex QCQP. \rev{Current} differentiable solvers for non-convex problems can return \rev{gradients of spurious local minima rather than the gradients of the global solution}, corrupting the learning process. In contrast, the gradients produced by the SDPRLayer \rev{correspond to the global solution as long as the relaxation is tight, leading to better training. Even when the relaxation is not tight, the relaxed, SDP solution is still provided, and can be used to find good initializations for local methods. The relaxed solution is also made differentiable} by leveraging existing differentiable convex optimization layers.}}
	\label{fig:abstract}
\end{figure}

Despite its recent successes, there is a fundamental issue with differentiable optimization that has gone unaddressed in the literature; the majority of optimization problems in robotics are non-convex and hence prone to convergence to local optima, rather than \rev{a} global optimum. In particular, if the optimization layer \rev{converges to spurious local minima}, the gradients that are backpropagated through the network will \rev{not correspond to the true global optimum}, \rev{and can therefore corrupt} the training/optimization process.\footnoterev{We acknowledge that, in some cases, finding \emph{any} stationary point may be sufficient for the objectives of the practitioner and the associated gradients may in fact be desirable. In this work, we take the stance that the goal of an optimization problem is to find the \emph{global} optimum and that convergence to local minima is undesirable.\label{ft:correct-grad}}
We will see that this can have serious consequences, leading not only to longer training times, but also outright failure to meet objectives of the training process. 

In this paper, we address this issue by \rev{taking inspiration from} a recent body of work in the robotics and vision communities that deals with so-called \emph{certifiably correct methods}~\cite[Section 2.2]{rosenAdvancesInferenceRepresentation2021} (also called \emph{certifiable algorithms}~\cite{yangCertifiablyOptimalOutlierRobust2023}). These methods use convex semidefinite relaxations of non-convex, polynomial optimization problems (POPs) to either directly find a global optimum or provide a certificate of global optimality for a given solution. \rev{Certifiably correct methods originate in the robotics community, but are closely related to the Moment-SOS hierarchy, which seeks to solve POPs using an \emph{automated hierarchy} of SDP relaxations~\cite{henrionMomentSOSHierarchyLectures2021}. However, since the goal certifiably correct methods is application in robotics, they tend to focus on bespoke relaxations that are tailored towards \emph{efficiently} solving or certifying solutions.}


\rev{Similar to the Moment-SOS hierarchy,} our approach is to directly solve the semidefinite relaxations \rev{of robotics problems. When the solution to the relaxation can be directly used to obtain a globally optimal solution for the original POP, the relaxation is said to be \emph{tight}}.\footnote{The astute reader may argue that in cases where strong duality holds at the solution, the solution can be certified without directly solving the SDP (e.g., scalar-weighted pose graph optimization \cite{rosenSESyncCertifiablyCorrect2019,holmesEfficientGlobalOptimality2023}). In our experience, the majority of robotics problems require the addition of so-called \emph{redundant constraints} to `tighten' the relaxation (see below in Section~\ref{sec:tightening_bg} and Appendix~\ref{sec:address_tightness}). However, when these constraints are used, certifying a solution amounts to solving an equivalent (dual form) SDP~\cite{dumbgenGloballyOptimalState2024}.} \rev{Even when the relaxation solution is not tight, it can still provide useful information about the original problem (see Section~\ref{sec:recourse}).}

\subsection{Contributions}\label{sec:contribution}

We focus on providing \emph{differentiable, globally optimal solutions} to POPs. Although there are existing differentiable solvers that could potentially find and differentiate first-order critical points for this class of problems, the resulting derivatives \rev{may not correspond to the solution of the optimization problem (i.e., the global optimum)}. In general, such solvers are only guaranteed to find local optima.

As one of our key contributions, we provide our approach as an optimization layer, which we call the \textit{SDPRLayer}\footnote{The naming is based on CVXPYLayers~\cite{agrawalDifferentiableConvexOptimization2019} with SDPR standing for Semidefinite Programming Relaxation.}, that conveniently encapsulates the \rev{parameterized} POP. The layer is guaranteed to yield \textit{certifiably correct, differentiable} solutions whenever the solution can be retrieved from the convex relaxation (i.e., the relaxation is tight). \rev{We do this \emph{efficiently} by leveraging the optimality conditions of the original, non-convex problem. For non-tight relaxations, we also provide the solution to the SDP, which we differentiate via existing, implicit-differentiation techniques for convex problems.} 

\rev{At the core of our approach is a theoretical result that provides conditions under which solution gradients can both be certified and efficiently computed. We view this result as a second key technical contribution.} 

\rev{Our} layer can be easily embedded in end-to-end learning pipelines and bilevel optimization for robotics problems. As a third contribution, we use simulated and real-world examples to demonstrate the potential pitfalls of relying on local solvers in this context. Crucially, we demonstrate that if local minima are used for backpropagation, the resulting gradients can be quite \rev{different from those of the global minimum}. We also show that the SDPRLayer can be used for large-scale training of a robotics perception pipeline.

\subsection{Outline}

In the next section (Section~\ref{sec:related}), we review works that are closely related to this paper. 
In Section~\ref{sec:background}, we review \rev{notation and} relevant background material on implicit differentiation and semidefinite programming that may not be common knowledge to all readers, but that can be safely skipped by domain experts. Section~\ref{sec:theory} provides the theoretical result that underpins our methodology\rev{, while Section~\ref{sec:implementation} describes the implementation of our layer}. In Section~\ref{sec:experiments}, we provide two examples that demonstrate the advantages of our approach and a third that demonstrates its application to a real-world robotics problem. Finally, in Section~\ref{sec:conclusion}, we present our conclusions, highlight the limitations of the current approach and provide ideas for future directions.

\section{Related Work}\label{sec:related}

\subsection{Certifiably Correct Optimization in Robotics}\label{sec:cert-corr-opt}

As mentioned, we rely on advances in certifiably correct optimization methods, which provide guarantees on the global optimality of solutions. In robotics, this approach has been applied to robust state estimation~\cite{yangCertifiablyOptimalOutlierRobust2023,yangTEASERFastCertifiable2021}, sensor calibration~\cite{wiseCertifiablyOptimalMonocular2020,giamouConvexIterationDistanceGeometric2022}, inverse kinematics~\cite{giamouConvexIterationDistanceGeometric2022}, image segmentation~\cite{huAcceleratedInferenceMarkov2019}, rotation synchronization~\cite{dellaertShonanRotationAveraging2020,erikssonRotationAveragingStrong2018}, pose-graph optimization~\cite{rosenSESyncCertifiablyCorrect2019,barfootCertifiablyOptimalRotation2024}, multiple-point-set registration~\cite{chaudhuryGlobalRegistrationMultiple2015, iglesiasGlobalOptimalityPoint2020}, range-only localization~\cite{dumbgenSafeSmoothCertified2023,goudarOptimalInitializationStrategies2024}, and range-aided SLAM~\cite{papaliaCertifiablyCorrectRangeAided2023}, among others. It is also worth noting that a similar approach using Sums-of-Squares (SOS) polynomials has been applied in the non-linear control community for automatic synthesis or verification of Lyapunov functions~\cite{majumdarRecentScalabilityImprovements2020,kangVerificationSynthesisRobust2023}. \rev{We note that any problem to which certifiably correct methods can be applied (including all those stated above) can necessarily be formulated as a POP.}

\subsection{Differentiable Optimization in Robotics}

New tools geared towards differentiable optimization in a  robotics setting have been recently presented. Theseus, developed by~\citet{pineda2022theseus}, present\rev{ed} a differentiable optimization layer for unconstrained, non-linear least-squares problems that typically appear in robotics. Crucially, this tool buil\rev{t} off work by~\citet{teedTangentSpaceBackpropagation2021} to solve problems with Lie Group constraints. 
For trajectory optimization, CALIPSO provide\rev{d} differentiable solutions using an interior-point solver and \rev{was} able to handle conic and complementarity constraints, which often occur when dealing with friction and contact in robotic motion~\cite{howellCALIPSODifferentiableSolver2023a}. Many of these methods \emph{implicitly differentiate} the conditions of optimality, a technique which, to date, has proven to be the most accurate and efficient way to generate gradient information of an optimum~\cite{blondelEfficientModularImplicit2022}. 

Many roboticists have adopted differentiable optimization layers into their pipelines. Building off their seminal work in differentiating quadratic programs (QPs)~\cite{amosOptNetDifferentiableOptimization2021},~\citet{amosDifferentiableMPCEndtoend2018a} presented a differentiable Model Predictive Control (MPC) framework that enabled learning of dynamics and objective functions. This work was subsequently extended to integrate Reinforcement Learning (RL) and safe learning into MPC~\cite{romeroActorCriticModelPredictive2023,zanonSafeReinforcementLearning2021,eastInfinitehorizonDifferentiableModel2020}. 
In robot planning, several works have developed differentiable approaches to dynamic programming~\cite{jalletImplicitDifferentialDynamic2022,dinevDifferentiableOptimalControl2022,jinPontryaginDifferentiableProgramming2020}.
Differentiable optimization has been used in physics-based simulators to enable efficient training of RL control strategies~\cite{deavilabelbute-peresEndtoEndDifferentiablePhysics2018} or robotic hardware design~\cite{xuEndtoEndDifferentiableFramework2021}. 
In state estimation, DROID-SLAM, which introduced a differentiable, end-to-end, neural architecture for visual Simultaneous Localization and Mapping (SLAM)~\cite{teedDROIDSLAMDeepVisual}, is most notable, though there have been several works on differentiable SLAM~\cite{jatavallabhulaNablaSLAMDense2020,fuISLAMImperativeSLAM2023}. Differentiable optimization has also been used for factor-graph-based estimators~\cite{qadriLearningObservationModels2023} and smoothers~\cite{yiDifferentiableFactorGraph2021}, tactile sensing~\cite{sodhiLearningTactileModels2021}, and shape estimation~\cite{fanRevitalizingOptimization3D2021}.

The most closely related work to ours is SATNet, introduced by~\citet{wangSATNetBridgingDeep2019}, which use\rev{d} semidefinite relaxations to generate approximate, differentiable solutions to the maximum satisfiability problem in combinatorial optimization.~\citet{amosDifferentiableMPCEndtoend2018a} also \rev{took} a similar approach, applying a convex \emph{quadratic} approximation to a non-convex MPC problem. Our approach is a \rev{more} general framework directed at providing \emph{exact}, differentiable solutions to robotics problems.~\citet{talakCertifiableObjectPose2023} also \rev{made use of certifiably correct methods to verify correctness of solutions, but did not directly differentiate the optimization problem.}

\subsection{Implicit Differentiation}

A key element of modern differentiable optimization is so-called \emph{implicit differentiation}, which applies the \rev{Implicit Function Theorem (IFT)} to the set of conditions \rev{-- known as the Karush-Kuhn-Tucker (KKT) conditions --} that characterize \rev{a local} optimum~\cite{dontchevImplicitFunctionsSolution2014}. The main idea is that the \rev{zero-level} set of the KKT conditions defines an implicit function \rev{that uniquely relates the parameters to the primal solution. Under certain conditions, }this relationship is differentiable, providing a means to \rev{backpropagate gradient information} through optimization problems. Prior differentiation methods, such as unrolling, are either less accurate~\cite{blondelEfficientModularImplicit2022}, slower, or more memory intensive than implicit differentiation~\cite{pineda2022theseus}.

Implicit differentiation can be applied to \emph{any} local optimum satisfying the (second-order) Karush-Kuhn-Tucker (KKT) conditions~\cite{giorgiTutorialSensitivityStability}.  When the problem is convex, implicit differentiation becomes much more powerful since the KKT conditions are sufficient and necessary for global optimality~\cite{boydConvexOptimization2004}.\footnote{Subject to an appropriate regularity condition such as Slater's condition.\label{fn:constraint_qual}} As such, it is often applied to convex problems or convex approximations of non-convex problems~\cite{amosOptNetDifferentiableOptimization2021,deavilabelbute-peresEndtoEndDifferentiablePhysics2018,wangSATNetBridgingDeep2019}. \rev{One notable example is }CVXPYLayers~\cite{agrawalDifferentiableConvexOptimization2019}, a general differentiable solver for convex problems\rev{, which} interfaces the popular CVXPY parser~\cite{diamondCVXPYPythonEmbeddedModeling} with a differentiable conic optimization solver~\cite{agrawalDifferentiatingConeProgram2019} \rev{and} has become the de-facto standard for differentiating convex programs.

\section{Background}\label{sec:background}

\subsection{Notation}\label{Sec:Notn}
We denote matrices with bold-faced, capitalized letters, $ \bm{A} $, column vectors with bold-faced, lower-case letters, $ \bm{a} $, and scalar quantities with normal-faced font, $ a $.
Let $ \mathbb{S}^{\nvar} $ (resp. $ \mathbb{S}_+^{\nvar} $) denote the space of $ \nvar $-dimensional symmetric (resp. positive, semidefinite) matrices, and $\mathbb{S}_1^{\nvar} \subset \mathbb{S}_+^{\nvar} $ denote the set of positive, semidefinite, rank-1 matrices. We also use the conic notation $\bm{X} \succeq \bm{0}$ in place of $\bm{X} \in \mathbb{S}_+^{\nvar} $. 
Let $ \bm{I} $ denote the identity matrix, whose dimension will be clear from the context or otherwise specified.
Let $ \bm{0} $ denote the matrix with all-zero entries, whose dimension will be evident from the context.
\rev{
Let $ \bm{A}^{\dagger} $ denote the Moore-Penrose pseudoinverse of a given matrix $ \bm{A} $ and let $\ker(\bm{A})$ and $\img(\bm{A})$ be the kernel space and image space of $\bm{A}$, respectively.
Let $ \vect{\bm{A}} $ denote the vectorization (reshape) of a given matrix $ \bm{A} $.
Let $ \otimes $ denote the Kronecker product.
}
Let $\left[N\right] = \left\{1,\ldots, N\right\} \subset \mathbb{N}$ be the set of indexing integers.
Let $\inner{\bm{A}}{\bm{B}} = \mbox{tr}(\bm{A}^{\top} \bm{B})$ denote the Frobenius inner product of matrices $\bm{A}$ and $\bm{B}$. 
Let $ \|\cdot\|_F $ denote the Frobenius norm.
Let $\mbox{SO}(d)$ denote the $d$-dimensional special orthogonal group.

\subsection{Semidefinite Relaxations of Polynomial Optimization Problems}\label{sec:sdp-relaxations}

Many key problems in robotics can be expressed as polynomial optimization problems (POPs)\rev{, including all of the problems mentioned in Section~\ref{sec:cert-corr-opt}}. In this section, we review the well-known procedure for deriving convex, SDP relaxations of a standard form of POP. This procedure was pioneered by~\citet{shorQuadraticOptimizationProblems1987} and has become the cornerstone of \textit{certifiably correct methods} in robotics and computer vision. 

We consider POPs that are parameterized on a variable, $\bth\in\mathbb{R}^d$, and are formulated in the standard \textit{homogenized, quadratically constrained quadratic problem} (QCQP) form,
\begin{problem}{opt:QCQPh}
	\begin{array}{rl}
		\min\limits_{\bx} & \bx^{\top}\bm{Q}_{\bth}\bx \\
		\mbox{s.t.} & \bx^{\top}\bm{A}_{\bth i}\bx = 0,\quad\forall i \in \indset{\ncon},\\
		&\bx^{\top}\bm{A}_0\bx = 1,
	\end{array}
\end{problem}
\rev{where $\bx\in\mathbb{R}^{\nvar}$, and $\bm{Q}_{\bth}\in\mathbb{S}^{\nvar}$ and $\bm{A}_{\bth i}\in\mathbb{S}^{\nvar}$ denote the (parameterized) cost and constraint matrices, respectively. The} \textit{homogenizing constraint}, $\bx^{\top}\bm{A}_0\bx = 1$, \rev{is used} to ensure that the optimization variable is \textit{homogeneous} (i.e., $ x_i = 1$, for some $i\in1,\dots,\nvar$) and $ \bx\in\mathbb{R}^{\nvar}$. The exact form of the matrices, $\bm{Q}_{\bth}$ and $\bm{A}_{\bth i}$, and an explanation of this transformation can be found in~\citet{cifuentesLocalStabilitySemidefinite2022}. \rev{We note that it is always possible to formulate a POP in the standard form of Problem~\eqref{opt:QCQPh}.}
\rev{The \emph{parameterized feasible set} of this problem is given by
\begin{equation}
	\feasset = \{ \bx\in\mathbb{R}^{\nvar}\vert~\bx^{\top}\bm{A}_{\bth i}\bx = 0,~\forall i \in \indset{\ncon},~ \bx^{\top}\bm{A}_0\bx = 1\}.
\end{equation}
}

Problem~\eqref{opt:QCQPh} can be re-expressed in terms of semidefinite matrices by using the properties of the trace operator\footnote{The trace is implicit in the inner product over matrices (i.e., Frobenius inner product).} as follows:

\begin{problem}{opt:R1SDP}
	\begin{array}{rl}
		\min\limits_{\bm{X}} & \inner{\bm{Q}_{\bth}}{\bm{X}}\\
		s.t.&\inner{\bm{A}_{\bth i}}{\bm{X}}= 0, \quad \forall i \in \indset{\ncon},\\
		& \inner{\bm{A}_0}{\bm{X}} = 1,\\
		& \bm{X} \succeq \bm{0},\\
		& \mbox{rank}(\bm{X})=1,
	\end{array}
\end{problem}
where the last two constraints implicitly enforce the fact that $\bm{X} = \bx\bx^{\top}$. \rev{All} the non-convexity \rev{of Problem~\eqref{opt:R1SDP} is contained in the single non-convex rank} constraint. It follows that we can find a convex relaxation of Problem~\eqref{opt:R1SDP} by removing the rank constraint: 
\begin{problem}{opt:SDPR}
	\begin{array}{rl}
		\min\limits_{\bm{X}} & \inner{\bm{Q}_{\bth}}{\bm{X}}\\
		s.t.&\inner{\bm{A}_{\bth i}}{\bm{X}}= 0, \quad \forall i \in \indset{\ncon},\\
		& \inner{\bm{A}_0}{\bm{X}} = 1,\\
		& \bm{X} \succeq \bm{0}.\\
	\end{array}
\end{problem}
This relaxation -- known as Shor's relaxation -- has been well studied by the optimization community. 

\rev{It can be shown that the Lagrangian of Problem~\eqref{opt:QCQPh} is given by
\begin{equation}
	L(\bx,\blam,\bth) = \bx^{\top} \left(\bm{Q}_{\bth}+ \bm{A}_{0}\lambda_0 + \sum\limits_{i=1}^{\ncon}\bm{A}_{\bth i}\lambda_i\right)\bx - \lambda_0
\end{equation}
where $\blam^{\top}=\begin{bmatrix} \lambda_0& \lambda_1\cdots\lambda_{\ncon} \end{bmatrix}$ are the Lagrange multipliers associated with the constraints ($\lambda_0$ being associated with the homogenizing constraint).\footnoterev{See~\citet{boydConvexOptimization2004} for an overview on Lagrangian duality theory.} It is well known that Problem~\eqref{opt:QCQPh} and Problem~\eqref{opt:SDPR} share a common Lagrangian dual problem:
\begin{equation}
	\begin{array}{rl}
		\min\limits_{\bm{H},\blam} & \lambda_0\\
		s.t.&\bm{H} = \bm{Q}_{\bth}+ \bm{A}_{0}\lambda_0 + \sum\limits_{i=1}^{\ncon}\bm{A}_{\bth i}\lambda_i,\\
		& \bm{H} \succeq \bm{0},
	\end{array}
\end{equation}
 where $\bm{H}$ is the dual matrix associated with the positive semidefinite constraint on $\bm{X}$. In what follows, we will refer to $\bm{H}$ as the \emph{certificate matrix} associated with the solution, since constructing such a matrix can be used to certify global optimality of solutions~\cite{carloneLagrangianDuality3D2015}. 
}

\subsubsection{Recovering A Global Solution}\label{sec:soln-recovery}
When the optimal solution of Problem~\eqref{opt:SDPR}, $ \bm{X}^* $, satisfies $ \mbox{rank}\hspace*{-2pt}\left(\bm{X}^*\right)=1 $ then the convex relaxation is said to be \textit{tight} to the original QCQP and the SDP solution can be factorized as $ \bm{X}^* = \bx^* \bx^{*\top} $ to obtain the \textit{globally optimal} solution, $ \bx^* $, of Problem~\eqref{opt:QCQPh}. This result has been proved rigorously by~\citet{cifuentesLocalStabilitySemidefinite2022}, among others, and depends on the inherent properties of the QCQP formulation and the fact that strong duality holds \emph{generically} for SDPs.\footnote{Slater's condition, which holds quite generally for SDPs (and in all of our problems), guarantees that strong duality holds.} The stability of the \emph{tightness} of semidefinite relaxations under perturbations to the objectives and constraints of the original QCQP was also studied extensively by~\citet{cifuentesLocalStabilitySemidefinite2022}.~\rev{This property is important in our context, since it suggests that tuning the parameters of a problem with a tight relaxation will not cause tightness to be lost.}

\subsubsection{\rev{Redundant Constraints and Solving SDPs}}\label{sec:tightening_bg}

For some problems in robotics and vision, the SDP relaxation is not immediately tight, but can be made so by adding so-called \emph{redundant constraints} to the original QCQP (e.g., ~\cite{brynteTightnessSemidefiniteRelaxations2022,yangTEASERFastCertifiable2021}). Although redundant in the original QCQP, these constraints are \emph{not redundant} for the SDP relaxation. 
In our recent work\rev{s}, we have shown how these redundant constraints can be efficiently and automatically generated for use in different state estimation problems that do not initially have tight relaxations~\cite{dumbgenGloballyOptimalState2024,holmesSemidefiniteRelaxationsMatrixWeighted2024}.

More generally, it has been shown that, subject to a mild technical condition \cite{nieOptimalityConditionsFinite2014}, any POP (including those problems that we have cited in Section~\ref{sec:related}) can be tightened via \rev{ the \emph{Moment-SOS Hierarchy}}~\cite{henrionMomentSOSHierarchyLectures2021}.\footnote{We refer here to the fact, introduced by \citet{nieOptimalityConditionsFinite2014}, that Lasserre's hierarchy has finite convergence for \emph{generic} POPs, as long as the Archimedean condition holds (i.e.\rev{,} compactness of the feasible set)} \rev{Put simply, the} hierarchy involves the addition of redundant variables and constraints to the problem that tighten the SDP relaxation.~\citet{yangCertifiablyOptimalOutlierRobust2023} provide a thorough-yet-accessible introduction to these ideas for robotics practitioners.

In general, SDPs can be solved in \textit{polynomial time} using interior-point methods~\cite{vandenbergheSemidefiniteProgramming1996}. These methods are fast for small problems and even tractable for problems of up to thousands of variables, but become prohibitive when real-time deployment is desired. 
\rev{Recent advances leverage low-rank SDP techniques such as that of \emph{Burer and Monteiro}~\cite{burerNonlinearProgrammingAlgorithm2003a,boumalNonconvexBurerMonteiro2016} and the \emph{Riemannian Staircase}~\cite{rosenSESyncCertifiablyCorrect2019} to solve larger problem instances in real time. However, our focus is on problems that require redundant constraints, which, to date, require solving the SDP directly. 
}

\rev{
\subsection{Implicit Differentiation of Equality-Constrained Optimization Problems}

In this section, we briefly review the mechanics of implicit differentiation as they apply to equality-constrained optimization problems.
Our development is largely based on the tutorial by \citet{giorgiTutorialSensitivityStability}.

Consider the following optimization problem that is parameterized on $\bth\in\mathbb{R}^d$,
\begin{problem}{opt:imp-diff-ex}
	\begin{array}{rl}
		\min\limits_{\bx\in\mathbb{R}^n} & f(\bx, \bth)\\
		\mbox{s.t.} & \bm{g}(\bx,\bth) = \bm{0},
	\end{array}
\end{problem}
where both $f$ and $\bm{g}$ are smooth, differentiable functions and $\bm{g}:\mathbb{R}^{n}\times\mathbb{R}^d\rightarrow \mathbb{R}^m$ is a vector-valued function representing $m$ constraints. Letting $\blam\in\mathbb{R}^m$ be the Lagrange multiplier vector associated with the constraints, the Lagrangian of the optimization is given by
\begin{equation}
	L(\bx,\blam,\bth)= f(\bx, \bth) + \blam^{\top}\bm{g}(\bx,\bth).
\end{equation}
The \emph{first-order KKT conditions} for Problem~\eqref{opt:imp-diff-ex} can be succinctly described in terms of a zero-level set of a function:
\begin{equation}
	\bm{k}(\bx,\blam,\bth) = \begin{bmatrix}
	\nabla_{\bx}L(\bx,\blam,\bth) \\ \bm{g}(\bx,\bth)
	\end{bmatrix}=\bm{0}.
\end{equation}

For a given parameter value, $\bar{\bth}$, any \emph{primal-dual pair}, $(\bar{\bx}, \bar{\blam})$, corresponding to a \emph{local optimum} of Problem~\eqref{opt:imp-diff-ex} will necessarily satisfy these conditions. These conditions are also sufficient for \emph{strict} local optimality if, in addition, the pair also satisfies the so-called \emph{second-order sufficiency conditions} (SOSC), 
\begin{equation}
	\bm{v}^T\nabla_{\bx}^2L(\bx, \blam,\bth)\bm{v}>0, \quad \forall \bm{v} \in \ker\nabla_{\bx}\bm{g}(\bx,\bth).
\end{equation}

In this paper, we are mainly concerned with the \emph{set-valued solution mapping}, $\solnmap:\mathbb{R}^{\nparam}\rightrightarrows\mathbb{R}^{\nvar+\ncon}$,\footnoterev{We use two arrows, $\rightrightarrows$, to denote a set-valued mapping. That is, a mapping that acts on a set and produces another set. In our case, the input set (the parameter set) is always singleton.} from parameters to the \emph{set} of primal-dual solutions,

\begin{equation}\label{eqn:solnmap}
	\solnmap : \bth \mapsto \{\bz\in\mathbb{R}^{\nvar+\ncon} \vert~ \bm{k}(\bz,\bth)=\bm{0}\},
\end{equation}
where, for convenience, we have concatenated the primal-dual pair into a single vector, $\bz^{\top}=\begin{bmatrix}\bx^{\top}&\blam^{\top} \end{bmatrix}$. More importantly, we wish to understand the differential relationship between the input parameters and the solution. Adopting the notion of \emph{differentials} from \citet{magnusMatrixDifferentialCalculus2019}, we consider the following differential relationship,
 \begin{equation}\label{eqn:kkt-differential}
	 \dd\bm{k}(\bm{z},\bth) = \bm{M} \dd\bm{z} + \bm{N} \dd\bth = \bm{0},
 \end{equation}
where
\begin{equation}
	\bm{M} = \frac{\dd\bm{k}(\bm{z},\bth)}{\dd\bm{z}} = \begin{bmatrix}
		\nabla_{\bx}^2 L(\bm{z}, \bth) & \nabla_{\bx} g(\bx, \bth) \\ \nabla_{\bx} g(\bx, \bth)^{\top} & \bm{0}
	\end{bmatrix},
\end{equation}
and $\bm{N} = \frac{\dd\bm{k}(\bm{z},\bth)}{\dd\bth}$. Whenever the matrix $\bm{M}$ is non-singular, we can apply the classical version of the IFT to $\bm{k}$.\footnoterev{For a formal definition of the IFT, see~\cite[Theorem 1B.1]{dontchevImplicitFunctionsSolution2014}.} Broadly speaking, the IFT guarantees that, at least in a neighbourhood containing $\bar{\bth}$, the solution mapping in~\eqref{eqn:solnmap} is \emph{single-valued}, and its exact Jacobian is given by
 \begin{equation}\label{eqn:soln-jac}
	 \nabla_{\bth}\solnmap(\bar{\bth}) = -\bm{M}^{-1}\bm{N}.
 \end{equation}

In the context of both bilevel optimization frameworks and neural network training, this Jacobian is used to find optimal search directions to minimize an overall loss function, $\ell(\bth)$. In general, parameters, $\bth$, are provided to an optimization `layer', which then solves the inner optimization in a forward pass. In the backward pass, the Jacobian of the solution is used to backpropagate gradient information to the parameters.
}

In theory, Problem~\eqref{opt:QCQPh} can be solved using any non-linear, non-convex optimization solver and its solution map can be differentiated by applying the implicit function theorem to the KKT conditions, as shown above~\cite{fiaccoSensitivityStabilityAnalysis1990,blondelEfficientModularImplicit2022}. 
However, in practice, there is a serious issue with this approach; there is no guarantee that the non-linear solver will converge to the global optimum of Problem~\eqref{opt:QCQPh}. Indeed, if the solver converges to a local optimum, then subsequent differentiation will occur with respect to the local solution instead of the true solution. 

\rev{
On the other hand, if global optimality of the solution can be certified, then gradients associated with that solution are also certified. Our approach is to directly solve the SDP relaxation in Problem~\eqref{opt:SDPR} in order to obtain a globally optimal solution, which often requires the use of redundant constraints to make the relaxation tight. Efficient differentiation of this solution is the subject of the next section.

\section{Differentiation of Certified Solutions}\label{sec:theory}

Given a tight solution to Problem~\eqref{opt:SDPR}, there are several ways that we could implicitly differentiate the solution. One approach would be to implicitly differentiate the SDP solution itself using the CVXPYLayers framework. This approach has the advantage of efficiently reusing the Lagrange multipliers and certificate matrix that are computed as a byproduct of solving the SDP. Unfortunately, tight SDP relaxations often violate the solution uniqueness assumptions of CVXPYLayers (and its underlying solver~\cite{agrawalDifferentiatingConeProgram2019}), meaning that the gradients they return cannot necessarily be trusted.\footnoterev{CVXPYLayers requires uniqueness of both the primal and the dual solution. It is known that low-rank SDPs are often primal degenerate, meaning that their dual solutions are not unique~\cite{yangCertifiablyOptimalOutlierRobust2023}. When the SDP solution has a rank of one, the results in~\cite{alizadehComplementarityNondegeneracySemidefinite1997} show that the dual is non-unique whenever $\ncon + 1 > \nvar$. This is quite often the case when we add redundant constraints to tighten the relaxation.} 

Another approach would be to apply implicit differentiation to the original QCQP at a globally optimal solution obtained from the relaxation. The KKT system of the original QCQP is much smaller than that of its SDP relaxation and is therefore more computationally efficient to solve. However, the existing theory requires the constraint gradients to be linearly independent, which is violated when redundant constraints are present.\footnoterev{See \cite{giorgiTutorialSensitivityStability} for an extensive survey of the existing theory of perturbation of optimization problems.} As a result, this approach would require us to first remove the redundant constraints and use the primal solution to recompute the Lagrange multipliers and certificate matrix before implicitly differentiating.

The goal of this section is to use an alternate version of the IFT that allows the use of redundant constraints and obviates the need for recomputation of the multipliers and certificate. We first show how the classic IFT can be applied to Problem~\eqref{opt:QCQPh} when no redundant constraints are present and highlight the reason that it fails when redundant constraints are required. We then introduce an alternate version of the IFT that can be applied even when redundant constraints are used and provide our main result: under certain conditions, we can compute the Jacobian in \eqref{eqn:soln-jac} using the pseudoinverse of a matrix related to $\bm{M}$.

In Section~\ref{sec:jac-compare}, we empirically show that the Jacobians of all the approaches discussed in this section are, to high precision, equal.

\subsection{Differentiation via Classic Implicit Function Theorem}

We first consider the situation in which no redundant constraints are needed to tight the relaxation and Problem~\eqref{opt:QCQPh} satisfies the Linear Independence Constraint Qualification (LICQ). That is, the gradients of the constraints are linearly independent. 

The KKT conditions associated with Problem~\eqref{opt:QCQPh} are given by
\begin{equation}\label{eqn:kkt-conditions}
		\bm{k}(\bm{z}, \bth) = \begin{bmatrix}
	2\bm{H}(\blam, \bth)\bx \\ \bm{g}(\bx,\bth)
	\end{bmatrix},~\bm{g}(\bx,\bth) = \begin{bmatrix}
		\bx^{\top}\bm{A}_{\bth 1}\bx \\ \vdots \\ \bx^{\top}\bm{A}_{\bth \ncon}\bx \\ \bx^{\top}\bm{A}_0\bx-1
	\end{bmatrix},
\end{equation} 
where $\nabla_{\bx}^2 L(\bm{z}, \bth) = 2\bm{H}(\blam, \bth)$ is exactly the certificate matrix described in Section~\ref{sec:sdp-relaxations}. We hereafter refer to $\bm{H}(\blam, \bth)$ as $\bm{H}$, dropping dependency for the sake of brevity. The solution map corresponding to these KKT conditions is
\begin{equation}\label{eqn:solnmap-qcqp}
	\solnmap: \bth \mapsto \{\bz\in\mathbb{R}^{\nvar+\ncon+1} \vert~ 2\bm{H}\bx = \bm{0},~\bm{g}(\bx,\bth)=\bm{0}\}.
\end{equation}
Considering the differential relationship in~\eqref{eqn:kkt-differential}, we have
\begin{equation}
	\bm{M} = 2\begin{bmatrix} \bm{H} & \bm{G}^{\top} \\ \bm{G} & \bm{0} \end{bmatrix},
\end{equation}
where the rows of $\bm{G}\in \mathbb{R}^{\ncon+1\times\nvar}$ are the constraint gradients,
\begin{equation}\label{eqn:contraint-grad-mat}
	\bm{G} = \nabla_{\bx}\bm{g}(\bx,\bth) = \begin{bmatrix}\bm{A}_{\bth 1}\bx & \cdots & \bm{A}_{\bth \ncon}\bx & \bm{A}_0\bx \end{bmatrix}^{\top}.
\end{equation}
An explicit definition of $\bm{N}$ is deferred to Appendix~\ref{sec:kkt-jac-params} since it is not required for our main discussion at this point. 

It can be shown that when the primal-dual solution satisfies the SOSC, $\bm{M}$ is invertible and the Jacobian of the solution map,~\eqref{eqn:solnmap-qcqp}, is exactly given by \eqref{eqn:soln-jac} (see~\cite{fiaccoNonlinearProgrammingSequential1990}).

When we need to introduce redundant constraints to tighten the SDP relaxation of Problem~\eqref{opt:QCQPh}, we can no longer apply the classical IFT; the fact that some constraints are redundant implies that $\bm{G}$ has linearly dependent columns and that $\bm{M}$ is singular. Equivalently, there are infinitely many solutions for the Lagrange multipliers and $\solnmap(\bth)$ is \emph{necessarily set-valued}. However, the next section shows that, under certain conditions, we can still recover an exact Jacobian, even when redundant constraints are used.

\subsection{Differentiation With Implicit Selections}\label{sec:imp-sel}

In solving Problem~\eqref{opt:QCQPh} for a particular parameter, $\bar{\bth}$, we \emph{select} a particular primal-dual solution from the set, $\bar{\bm{z}} \in \solnmap(\bth)$, which certifies global optimality. Even when $\bm{M}$ is does not have full column rank, there is a version of the implicit function theorem that still allows us to differentiate our selected solution by inferring the existence of single-valued function that is contained within the set-valued solution map. This concept of a \emph{local selection} function is borrowed from functional analysis and is formally defined as follows~\cite{dontchevImplicitFunctionsSolution2014}:

\begin{definition}[Local Selection]
	Given a set-valued mapping $\solnmap:\mathbb{R}^{\nparam}\rightrightarrows\mathbb{R}^{\nvar+\ncon+1}$ and a pair, $(\bar{\bm{z}}, \bar{\bth})$, such that $\bar{\bm{z}}\in \solnmap(\bar{\bth})$, a function $\bm{w}:\mathbb{R}^{\nparam}\rightarrow\mathbb{R}^{\nvar+\ncon+1}$ is said to be a \emph{local selection} of $\solnmap$ around $\bar{\bth}$ for $\bar{\bm{z}}$ if $\bm{w}(\bar{\bth})=\bar{\bm{z}}$ and, for a neighbourhood $\mathcal{V}\subseteq\mathbb{R}^{\nparam}$ containing $\bar{\bth}$, we have that $\bm{w}(\bth) \in \solnmap(\bth)$ for all $\bth\in\mathcal{V}$.
\end{definition}

The local selection function has a well-defined Jacobian, as shown by the next theorem, adapted from~\cite[Exercise 1F.9]{dontchevImplicitFunctionsSolution2014}:

\begin{theorem}[Implicit Selections]\label{thm:imp-sel}
	Consider a function $ \bm{k} : \mathbb{R}^d \times \mathbb{R}^l \to \mathbb{R}^p $, where $p \leq l$, along with the associated solution mapping
	\begin{equation}
		\solnmap: \bth \mapsto \left\{ \bm{z} \in \mathbb{R}^l \mid \bm{k}(\bm{z}, \bth) = \bm{0} \right\} \quad \text{for } \bth \in \mathbb{R}^d.
	\end{equation}
	Let $\bm{k}(\bar{\bm{z}}, \bar{\bth}) = \bm{0}$, so that $\bar{\bm{z}} \in \solnmap(\bar{\bth})$. Assume that $\bm{k}$ is strictly differentiable at $(\bar{\bm{z}}, \bar{\bth})$ and suppose further that the partial Jacobian  $\nabla_{\bm{z}} \bm{k}(\bar{\bm{z}}, \bar{\bth})$ is of rank $m+1$. Then the mapping $\solnmap$ has a local selection $\bm{w}$ around $\bar{\bth}$ for $ \bar{\bm{z}}$ that is strictly differentiable at $\bar{\bth}$ with Jacobian
	\begin{equation}\label{eqn:imp-sel-grad}
		\nabla \bm{w}(\bar{\bth}) = \bm{M}^\top \left( \bm{M} \bm{M}^\top \right)^{-1} \bm{N},
	\end{equation}
	where $\bm{M} = \nabla_{\bm{z}} \bm{k}(\bar{\bm{z}}, \bar{\bth})$ and $\bm{N} = \nabla_{\bth} \bm{k}(\bar{\bm{z}}, \bar{\bth})$.
\end{theorem}
\begin{remark}[Strict Differentiability]
	The notion of \emph{strict differentiability} at a given point is rigorously defined in \citet{dontchevImplicitFunctionsSolution2014} and is equivalent to continuous differentiability at every point in an open set containing the point (c.f. Exercise 1D.8 in~\cite{dontchevImplicitFunctionsSolution2014}). For our purposes, this distinction is tautalogical, since we will consider continuously differentiable functions.
\end{remark}

Given a \emph{selected} solution for which global optimality is guaranteed,~\eqref{eqn:imp-sel-grad} describes how this selection will change as the input parameters are locally perturbed. It is perhaps not surprising that the gradient takes the form of a (right) Moore-Penrose pseudoinverse. However, we cannot apply this result directly, since, in our case, the solution map may not have full row rank.

Our main result will require the following constraint qualification, which is weaker than the typical qualifications used in the optimization literature\footnoterev{LICQ or the Mangasarian-Fromovitz Constraint Qualification (MFCQ) are commonly used in sensitivity analysis of optimization problems~\cite{giorgiTutorialSensitivityStability}. Both of these qualifications imply the (ACQ), though the converse is not generally true.} and allows us to apply redundant constraints.

\begin{definition}[Abadie Constraint Qualification (ACQ)]
	Given a set of constraints $\bm{g} : \mathbb{R}^{\nvar} \to \mathbb{R}^{\ncon}$, let $\bm{\Omega} := \{ \bx \in \mathbb{R}^{\nvar} : \bm{g}(\bx) = \bm{0} \}$ be the feasible set.
	The \emph{Abadie constraint qualification} (ACQ) holds at $\bx \in \bm{\Omega}$, if $\bm{\Omega}$ is a smooth manifold nearby $\bx$, and 
	\begin{equation}
	\rank(\nabla \bm{g}(x)) = \codim_{\bx}(\bm{\Omega}),
	\end{equation} 
	where $\codim_{\bx}(\bm{\Omega}) = \nvar - \dim_{\bx}\bm{\Omega}$ denotes the local codimension of $\bm{\Omega}$ at $\bx$, $\dim_{\bx}\bm{\Omega}$ denotes the local dimension of $\bm{\Omega}$ at $\bx$, and $\nabla \bm{g}$ denotes the Jacobian matrix.
\end{definition}
The ACQ holds in many robotics problems, especially in state estimation, where optimization variables such as rotations or poses are confined to a smooth manifold. 

We now provide our main theorem, which shows how the Jacobian of the solution can be obtained even when the problem has redundant constraints. 

\begin{theorem}[QCQP Jacobian]\label{thm:soln-jac}
	For some parameter, $\bar{\bth}$, let $(\bar{\bx},\bar{\blam})$ be a primal-dual solution satisfying the (first-order) KKT conditions for Problem~\eqref{opt:QCQPh} and let $\bar{\bm{H}}$ be the associated certificate matrix.
	Assume that the following conditions hold:
	\begin{enumerate}
		\item The set of parameterized matrices, $\{\bm{Q}_{\bth}, \bm{A}_{\bth i}\}$, are continuously differentiable with respect to $\bth$.
		\item The ACQ and SOSC hold at $\bar{\bx}$. \label{assmp:acq}
		\item The mapping from the parameters, $\bth$, to the feasible set, $\feasset$, is smooth near $\bar{\bth}$. \label{assmp:smooth-feas}
		\item The certificate matrix is positive semidefinite, $\bar{\bm{H}}\succeq\bm{0}$.
	\end{enumerate}

	\noindent Then $\bar{\bx}$ is the globally optimal solution for Problem~\eqref{opt:QCQPh} at $\bar{\bth}$ and its unique Jacobian with respect to $\bth$ is given by
	\begin{equation}\label{eqn:soln-jac-general}
		\bm{J} = -\bm{P}\bm{M}_r^{\dagger}\bm{N},
	\end{equation}
	where
	\begin{equation}\label{eqn:kkt-jac-red}
			\bm{P} = \begin{bmatrix}
			\bm{I} \\ \bm{0}
			\end{bmatrix},\quad\bm{M}_r = 2\begin{bmatrix} \bar{\bm{H}} & \bm{G}^{\top} \\ \bm{G}_r & \bm{0} \end{bmatrix}, 
	\end{equation}
	$\bm{M}_r^{\dagger} =\bm{M}_r^{\top}(\bm{M}_r\bm{M}_r^{\top})^{-1}$ denotes the (right) Moore-Penrose pseudoinverse, and $\bm{G}_r$ is obtained from $\bm{G}$ by removing any linearly dependent rows. 
\end{theorem}

\begin{IEEEproof}
	Our approach to demonstrate differentiability is inspired by Appendix A of~\cite{cifuentesLocalStabilitySemidefinite2022} and involves applying Theorem~\ref{thm:imp-sel} to the KKT conditions of Problem~\eqref{opt:QCQPh}. However, to apply the theorem we must first modify the KKT conditions so that $\bm{M}$ has full row rank.

	\def\nconstred{r}
	Let $\bm{g}_r(\bx,\bth)$ be a maximal subset of the constraint equations in $\bm{g}(\bx,\bth)$ such that the rows of the corresponding Jacobian, $\bm{G}_r \in \mathbb{R}^{\nconstred\times\nvar}$, are linearly independent, where\footnoterev{Note that here we have reordered the constraints such that the first $r$ are linearly independent.}
	\begin{gather}\label{eqn:cons-grad-reduced}
		\bm{G}_r = \nabla_{\bx}\bm{g}_r(\bx,\bth) = \begin{bmatrix}\bm{A}_{\bth 1}\bx & \cdots & \bm{A}_{\bth \nconstred}\bx \end{bmatrix}^{\top}.
	\end{gather}
	Under assumptions \ref{assmp:acq} and \ref{assmp:smooth-feas} in the theorem statement, Lemma A.8 of~\cite{cifuentesLocalStabilitySemidefinite2022} guarantees that the feasible set induced by $\bm{g}_r(\bx,\bth)$ is locally equivalent to the feasible set of Problem~\eqref{opt:QCQPh}, $\feasset$. Our solution mapping can therefore be re-expressed as follows:
	\begin{equation}
		\solnmap:~\bth\mapsto\{\bm{z}\in\mathbb{R}^{\nvar+\nconstred}\vert ~ \bar{\bm{H}}\bx=\bm{0},~\bm{g}_r(\bx,\bth)=\bm{0}\}.
	\end{equation}
	The Jacobian of this mapping with respect to $\bm{z}$ is exactly $\bm{M}_r$, as given in~\eqref{eqn:kkt-jac-red}. 
	
	We now show that $\bm{M}_r$ has linearly independent rows. Let $\bm{t}^{\top}=\begin{bmatrix} \bm{v}^{\top}& \bm{u}^{\top}\end{bmatrix} $ be such that $\bm{M}_r^{\top}\bm{t} = \bm{0}$. We have:
	\begin{gather*}
		\bar{\bm{H}} \bm{v} + \bm{G}_r^{\top}\bm{u} = \bm{0}, \\
		\bm{G}\bm{v} = \bm{0}.
	\end{gather*}
	The second line implies that $\bm{v}\in \ker{\bm{G}}=\ker{\bm{G}_r}$. Multiplying the first equation by $\bm{v}^{\top}$, we have
	\begin{equation*}
		\bm{v}^{\top}\bar{\bm{H}} \bm{v} + \bm{v}^{\top}\bm{G}^{\top}\bm{u} =\bm{v}^{\top}\bar{\bm{H}} \bm{v}= \bm{0},
	\end{equation*}
	since $\bm{v}\in \ker{\bm{G}}$. The SOSC assumption implies that $\bm{v}^{\top}\bar{\bm{H}} \bm{v}>0$ for all non-zero $\bm{v}\in \ker{\bm{G}}$. Combined with the equation above, we must have that $\bm{v} = \bm{0}$. 
	
	It follows that $\bm{u} = \bm{0}$ since $\bm{G}_r^{\top}\bm{u}=\bm{0}$ and $\bm{G}_r^{\top}$ has linearly independent columns by construction. Thus, the only vector $\bm{t}\in\ker{\bm{M}_r^{\top}}$ is the zero vector, so $\bm{M}_r$ has linearly independent rows.
	
	Continuous differentiability of the KKT conditions with respect to $\bm{z}$ and $\bth$ holds by the quadratic nature of the parameterized input matrices and by assumption, respectively. 

	All the conditions of Theorem~\ref{thm:imp-sel} are satisfied and, applying the theorem, there exists a local selection $\bm{w}_1$ of $\solnmap$ around $\bar{\bth}$ for $\bar{\bm{z}}$. Defining $ \begin{bmatrix} \bx_1^{\top} & \blam_1^{\top} \end{bmatrix} = \bm{w}^T_1(\bth)$, we have that the Jacobian of $\bx_1$ with respect to $\bth$ is given exactly by~\eqref{eqn:soln-jac-general}. 
	
	It remains to show that this Jacobian is, in fact, unique and not specific to the local selection $\bm{w}_1$. Let $\bm{w}_2$ be any other local selection of $\solnmap$ around $\bar{\bth}$ for $\bar{\bm{z}}$ and let $\begin{bmatrix} \bx_2^{\top} & \blam_2^{\top} \end{bmatrix} = \bm{w}^T_2(\bth)$. Since $\bm{w}_1(\bth), \bm{w}_2(\bth)\in\solnmap$ for all $\bth$ in a neighbourhood about $\bar{\bth}$, both selections satisfy the differential relationship given in~\eqref{eqn:kkt-differential}. Subtracting, it follows that,
	\begin{equation}
		\bm{M}_r\begin{bmatrix}
			d\bx_1-d\bx_2 \\ d\blam_1-d\blam_2
		\end{bmatrix}=\bm{0}
	\end{equation}
	Let $\bm{v} = d\bx_1-d\bx_2$ and $\bm{u} = d\blam_1-d\blam_2$. Then $\bm{v} \in \ker{\bm{G}}$ and $\bar{\bm{H}} \bm{v} + \bm{G}^{\top}\bm{u} = \bm{0}$. Multiplying by $\bm{v}^{\top}$, we have that $\bm{v}^{\top}\bar{\bm{H}} \bm{v} = 0$, which implies that $\bm{v}=\bm{0}$ (since the SOSC holds). Therefore, it must be that  $d\bx_1=d\bx_2$. It follows that the differential $d\bx_1$ is unique and has a unique Jacobian with respect to $\bth$ given by~\eqref{eqn:soln-jac-general}.

	Finally, it is well known that $\bar{\bm{H}}\succeq\bm{0}$ is sufficient to guarantee global optimality of $\bar{\bx}$~\cite{cifuentesLocalStabilitySemidefinite2022}. By extension,~\eqref{eqn:soln-jac-general} is the Jacobian of a globally optimal minimum.
\end{IEEEproof}

\begin{remark}[Smooth Feasible Set Assumption]
	The assumption of a smooth mapping from the parameters to the feasible set is more formally defined in Definition 4.1 of~\cite{cifuentesLocalStabilitySemidefinite2022}, but is akin to assuming that the set, $\mathcal{W}=\{(\bth,\bx)\vert~\bx\in\feasset\}$, is (locally) a smooth manifold near $(\bar{\bth},\bar{\bx})$. In many robotics contexts, only the cost of the optimization is a function of the parameters and this assumption always holds. This is the case in all our experiments.
\end{remark}

In our context, $\bar{\bx}$ in Theorem~\ref{thm:soln-jac} is obtained from a rank-one solution to the SDP relaxation, Problem~\eqref{opt:SDPR}. As mentioned above, this theorem provides a means by which we can reuse the certificate matrix and Lagrange multipliers (obtained as a byproduct when solving the SDP), to compute the Jacobian of the solution with respect to the parameters. Note that when no redundant constraints have been used to tighten the problem, Theorem~\ref{thm:soln-jac} still applies with $\bm{M}_r = \bm{M}$.

\subsection{Corank of the Certificate Matrix}\label{sec:corank}

The corank of the certificate matrix is important in our context for two key reasons. First, it has been shown that, when the corank of the certificate is one, the tightness of SDP relaxations is stable under small perturbations of the cost and constraint matrices~\cite{cifuentesLocalStabilitySemidefinite2022}. This provides us with some guarantee that, as long as we start with a tight relaxation, tuning the input parameters will not cause the tightness of the SDP to be lost.

Second, the following Lemma (adapted from \cite{cifuentesLocalStabilitySemidefinite2022}) shows that the same condition on the corank implies that the SOSC holds for our problem.

\begin{lemma}\label{thm:cert-positive}
	Let $\bm{H}$ be the certificate matrix associated with a solution $\bar{\bx}$ to Problem~\eqref{opt:QCQPh} and suppose that $\bar{\bm{H}}\succeq\bm{0}$ and $\corank{\bar{\bm{H}}}=1$. Then $\bm{v}^{\top}\bar{\bm{H}}\bm{v} > 0$ for all non-zero $\bm{v}$ such that $\bm{v}\in\ker{\bm{G}}$.
\end{lemma}
\begin{IEEEproof}
	Let $\bm{v}\in\ker{\bm{G}}$. The assumption $\bm{v}^{\top}\bar{\bm{H}}\bm{v} \geq 0$ holds with equality \emph{only if} $\bm{v}=\mu\bar{\bx}$ ($\mu\geq0$), since $\corank{\bar{\bm{H}}}=1$ and $\bar{\bm{H}}\bar{\bx}=\bm{0}$ imply that $\ker{\bar{\bm{H}}} = \img{\bar{\bx}}$. Since $\bm{v}\in\ker{\bm{G}}$ and $\bar{\bx}^{\top}\bm{A}_0$ is a row of $\bm{G}$, we have that $\bar{\bx}^{\top}\bm{A}_0 \bm{v}=\bm{v}_h=0$, where $\bm{v}_h$ is the element of $\bm{v}$ that corresponds to the homegenization index. Since $\bx_h=1$, $\bm{v}^{\top}\bar{\bm{H}}\bm{v} = 0$ only if $\mu=0$. Thus, $\bm{v}^{\top}\bar{\bm{H}}\bm{v} > 0$ whenever $\bm{v}\neq\bm{0}$.
\end{IEEEproof}
 
When \emph{strict complementarity} of the relaxation holds, the certificate corank condition above is equivalent to having a rank-one solution to the relaxation.
Strict complementarity has been shown to be \emph{generic} for SDPs~\cite{alizadehComplementarityNondegeneracySemidefinite1997} and, in practice, we have observed that many relaxations of robotics problems enjoy this property (c.f.~\cite{holmesSemidefiniteRelaxationsMatrixWeighted2024}). Indeed, this property holds for all the problems investigated in Section~\ref{sec:experiments}.

\section{Implementation}\label{sec:implementation}

In this section, we introduce our implementation, the SDPRLayer, which computes the differentiable, globally optimal solution to a polynomial optimization problem and can be embedded in any PyTorch autodifferentiation graph. As one of our contributions, we provide an encapsulation of our method in a PyTorch neural network module.

In a nutshell, this module takes as input a set of PyTorch tensors representing the parameterized cost and constraint matrices, $\{\bm{Q}_{\bth}, \bm{A}_{\bth i}\}$, and returns the solution to the optimization as (differentiable) PyTorch tensor, which can then be used in succeeding layers of the PyTorch compute graph. If redundant constraints are used to tighten the problem, we assume that the user provides a list of these constraints. 

As with many differentiable frameworks, the key components of our layer can be described in terms of a `forward' and `backward' function. The forward function computes the solution to the optimization problem and caches the primal-dual solution, while the backward function uses the cached solution to compute gradients of input parameters (tensors) from a known gradient of the solution. We provide details on the specific implementation of these functions in the remainder of this section. All operations are designed to be compatible with batched inputs and outputs.

\subsection{SDPRLayer Forward Function}

As mentioned, the forward function is used to find the primal-dual solution based on the parameterized input cost and constraint matrices.\footnoterev{Since the homogenizing constraint is \emph{always} required in our framework, we add it automatically. It does not need to be specified by the user.} By default, the primal-dual solution is found by formulating the SDP relaxation as a Disciplined Convex Program (DCP) and passing it to (a modified version of) CVXPYLayers~\cite{agrawalDifferentiableConvexOptimization2019}. Alternatively, the user can specify to solve the problem using Mosek or provide their own primal-dual solution, computed by a custom solver. The primal-dual solution is then cached for use by the backward function.

The forward function always returns the matrix solution of the SDP relaxation (Problem~\eqref{opt:SDPR}) and, \emph{when the relaxation is tight}, also returns the globally optimal solution vector to the non-convex problem (Problem~\eqref{opt:QCQPh}). Both the matrix and vector solutions are returned as differentiable PyTorch tensors.

\subsection{SDPRLayer Backward Function}

The backward function has been implemented for both the globally optimal vector solution to Problem~\eqref{opt:QCQPh} (when available) and the matrix solution to its relaxation, Problem~\eqref{opt:SDPR}. In practice, the input parameters of the SDPRLayer are the (vectorized) input matrices,
\begin{equation}\label{eqn:vect-input-mats}
	\bm{\nu}^{\top}=\begin{bmatrix}\vect{\bm{Q}_{\bth}}^{\top}& \vect{\bm{A}_{\bth,1}}^{\top}&\cdots&\vect{\bm{A}_{\bth,\ncon}}^{\top}\end{bmatrix}.
\end{equation} 
The interpretation is that $\bth$ is a set of parameters in the layers preceding the SDPRLayer, on which the cost and constraint matrices are parameterized. 

\def\grad#1{\nabla_{#1}\ell^{\top}} 
The PyTorch framework that we have adopted uses reverse-mode autodifferentiation, in which the adjoint of the Jacobian is used to backpropagate gradient information. Given the gradient of the loss function with respect to the solution, $\grad{\bx}$ (or $\grad{\bm{X}}$ if the matrix solution is used), the backward function computes the gradient of the loss with respect to $\bm{\nu}$,
\begin{equation}
	\grad{\bm{\nu}} = \bm{J}^{\top}\grad{\bx},
\end{equation}
where $\bm{J}$ is the solution map Jacobian that was discussed throughout Section~\ref{sec:theory}. The remainder of this section discusses the computation of this gradient in different cases. 

\subsubsection{Backpropagation via Implicit Selections (SDPR-IS)}\label{sec:backprop-sift}

When the solution is tight (i.e., the rank of the SDP is one), we can backpropagate gradients using the Jacobian in Theorem~\ref{thm:soln-jac}, regardless of whether or not redundant constraints are used (referred to as SDPR-IS hereafter).
We have,
\begin{equation}
	\grad{\bm{\nu}} =-\bm{N}^{\top}\bm{M}_r^{\top\dagger}\grad{\bx},
\end{equation}
For efficiency, we avoid explicit computation of the Jacobian (or its pseudoinverse) when computing the gradients. To do so, we first define an intermediate gradient, $\grad{\bm{y}}=\bm{M}_r^{\top\dagger}\grad{\bx}$, representing the gradient with respect to the KKT conditions. Noting that $\bm{M}_r^{\top}$ has full column rank, by the properties of the pseudoinverse we have 
\begin{equation}\label{eqn:grad-lstsqr}
	\grad{\bm{y}} = \arg\min\limits_{\bm{y}} \Vert\bm{M}_r^{\top}\bm{y} - \grad{\bx}\Vert_2^2.
\end{equation}

Similar to~\cite{agrawalDifferentiatingConeProgram2019}, we solve this optimization using the LSQR algorithm, which is specialized for sparse, unsymmetric linear systems~\cite{paigeLSQRAlgorithmSparse1982}. This algorithm effectively leverages the sparsity of $\bm{M}_r$ by only requiring matrix-vector products and is also robust to ill-conditioned problems, which is advantageous in situations where constraint gradients are \emph{nearly} dependent.\footnoterev{In practice, we have observed that the LSQR algorithm even provides meaningful gradients when some of the constraints are exactly dependent (i.e., $\bm{G}_r$ has linearly dependent rows).}  Gradients can then be computed via $\grad{\bm{\nu}}=-\bm{N}^{\top}\grad{\bm{y}}$, where $\bm{N}$ is as shown in Appendix~\ref{sec:kkt-jac-params}. 

Under the conditions of Theorem~\ref{thm:soln-jac}, the approach shown above is guaranteed to provide the gradients of the globally optimal solution with respect to the input parameters.

\subsubsection{Backpropagation via Classic IFT (SDPR-CIFT)}\label{sec:backprop-cift}

The method in the previous section represents the default behaviour of SDPRLayers, but we also provide an option to use the classic IFT to differentiate the original QCQP (referred to as SDPR-CIFT hereafter). In this case, the primal solution, $\bx$, is used to compute the Lagrange multipliers of a non-redundant subset of the constraints:
\begin{equation}
	\blam = -\bm{G}_r^{\top\dagger}\bm{Q}_{\bth}\bx,
\end{equation}
where the rows of $\bm{G}_r$ are the gradients of the non-redundant constraints, as in~\eqref{eqn:cons-grad-reduced}. Using the multipliers, we construct the matrices,
\begin{equation}
	\bm{M} = 2\begin{bmatrix} \bm{H}_r & \bm{G}_r^{\top} \\ \bm{G}_r & \bm{0} \end{bmatrix},\quad \bm{H}_r = \bm{Q}_{\bth}+ \bm{A}_{0}\lambda_0 + \sum\limits_{i=1}^{r}\bm{A}_{\bth i}\lambda_i.
\end{equation}
Finally, backpropagation is performed by solving the linear system,
\begin{equation}\label{eqn:kkt-lin-sys-cift}
	\bm{M}\bm{y} = \grad{\bx},
\end{equation}
and computing the gradients via $\grad{\bm{\nu}}=-\bm{N}^{\top}\bm{y}$. In practice, we also use the LSQR algorithm to solve~\eqref{eqn:kkt-lin-sys-cift}.

\subsubsection{Backpropagation via SDP Solution (SDPR-SDP)}\label{sec:backprop-sdp}

As mentioned, the (matrix) solution to the SDP relaxation is provided to the user, regardless of whether the solution is tight. We use CVXPYLayers, which implicitly differentiates the KKT conditions of the SDP relaxation rather than the QCQP, to compute the gradients~\cite{agrawalDifferentiableConvexOptimization2019} (referred to as SDPR-SDP hereafter). 

As mentioned in Section~\ref{sec:theory}, the requisite assumption of a unique primal-dual solution in CVXPYLayers is often violated in our context.\footnoterev{In particular, it is violated for the localization experiments in Sections~\ref{sec:stereo_exp} and \ref{sec:deeplearn}, though not in polynomial example in Section~\ref{sec:poly_exp}}
Despite this technical issue, we have empirically observed that backpropagation through the SDP relaxation yields the same gradients as the backpropagation through the QCQP, at least when the relaxation is tight (see Section~\ref{sec:jac-compare}).\footnoterev{Internally, CVXPYLayers uses a least-squares formulation to solve for the gradients during backpropagation, which provides a solution even when the KKT Jacobian (equivalent to $\bm{M}$ above) is not invertible. We suspect that the reason that the gradients are correct is that they correspond to a local selection, similar to our development above. However, further exploration of this idea is left as future work.} However, backpropagation through the SDP relaxation typically involves solving a much larger linear system\footnote{The number of primal and slack variables in the relaxation scales as $O(n^2)$.}, which is slower than differentiating through the non-convex problem when the problem is large.

\subsection{Recourse For Non-Tight Relaxations}\label{sec:recourse}

The SDP matrix solution, $\bm{X}$, can still be used to obtain a near-optimal solution, $\hat{\bx}$, by \emph{rounding} to the nearest feasible point for the original problem~(see, for example, \cite{rosenSESyncCertifiablyCorrect2019}). This rounding procedure is problem dependent, but typically involves differentiable operations such as singular value decomposition (SVD) and projection. 

The \emph{suboptimality gap} of the near-optimal solution can be computed as follows:
\begin{equation}
	\mu(\hat{\bx},\bm{X}) = \frac{\inner{\bm{Q}_{\bth}}{\hat{\bx}\hat{\bx}^{\top} - \bm{X}}}{\inner{\bm{Q}_{\bth}}{\bm{X}}}.
\end{equation}
The solution may be acceptable if this \emph{suboptimality gap} is low.
If the suboptimality gap is large, then the feasible point can still serve as an initial guess for a local solver, similar to the methodology used in~\cite{yangCertifiablyOptimalOutlierRobust2023}. There are then two possibilities for differentiation of the solution. If the local solver is also differentiable, then gradient information can be backpropagated through the local solver, rounding procedure, and CVXPYLayers implicit differentiation. Alternatively, the final solution can be passed back to the SDPRLayer and gradients can be computed via SDPR-CIFT (see Section~\ref{sec:backprop-cift}).

In this case, the theory does not guarantee correctness of the gradients of the solution. An interesting avenue of future work could include an investigation of the relationship between the suboptimality gap and the \emph{level} of correctness of the gradients. 

We have added tools to our implementation for assessing and improving tightness of a given relaxation. These tools, along with a general approach to tightening SDP relaxations, are described in Appendix~\ref{sec:address_tightness}.

}

\section{Experiments}\label{sec:experiments}

We now present a series of examples that demonstrate the utility of our method in comparison to (non-global) alternatives. The first two experiments are simulated and highlight the issues with naive application of \rev{local optimization methods in the context of} differentiable optimization. \rev{We use the second experiment to provide a detailed comparison of the Jacobians produced by the methods discussed above.} The final experiment shows that our approach can be used to train a deep neural network in a real-world robotics pipeline. \rev{In all experiments, the optimal solutions satisfy all the assumptions of Theorem~\ref{thm:soln-jac} and the corank of the certificate matrix was equal to one. We therefore use the SDPR-IS method in all cases, except when comparing Jacobians.} 

Note that our theory \rev{and implementation} allow both cost and constraints to be functions of parameters, but our examples focus on cases where the constraints are fixed.

\subsection{Polynomial Experiment}\label{sec:poly_exp}
\begin{figure*}
    \centering
    \includegraphics[width=\textwidth]{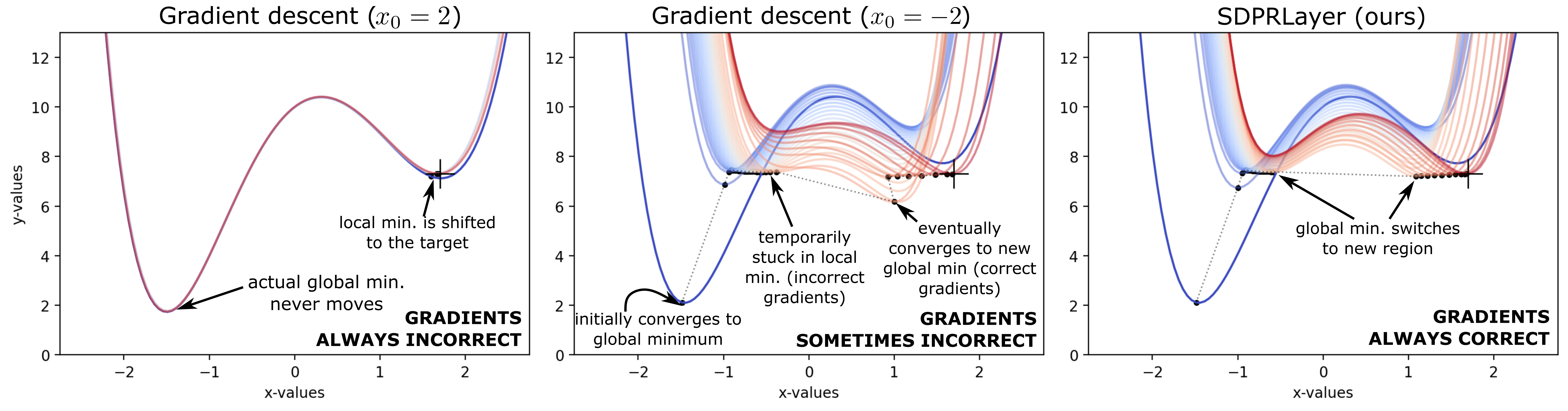}
    \caption{Evolution of the polynomial function throughout the bilevel optimization. Columns present different methods for solving the inner optimization problem: (\emph{left}) gradient descent initialized at $ x=2 $, (\emph{center}) gradient descent initialized at $ x=-2 $, and (\emph{right}) our approach. In all three cases, the colour of the function indicates the progress of outer loop iterations (blue at the beginning, red at the end), the black dots indicate the minima found by the inner optimization, and the plus sign indicates the target. All three cases converge, but gradient descent only converges to a valid solution when it is initialized well ($x_0=-2$) and, even then, is temporarily trapped in a local minimum. SPDRLayer converges to a valid solution, does not require initialization, and is even able to discontinuously `switch' the minimum to a new region.}
	\label{fig:poly_evo_ex}
\end{figure*}
In this section, we consider a \rev{toy example in} bilevel polynomial optimization that clearly illustrates the potential issues that can arise when local optimization is used \rev{and is assumed to converge to the global minimum}. The objective of this bilevel optimization problem is to find a sixth-order polynomial that has a global minimum at a pre-specified point, $(\bar{x}, \bar{y})$. The polynomial function is parameterized by its coefficients $\bm{\theta}$, 
\begin{equation}
	y(x,\bm{\theta}) = \sum_{i=0}^{6} \theta_i x^i.
\end{equation}
The task is split into an inner optimization, which attempts to find the global minimum of the current polynomial, and an outer optimization, which tunes the polynomial coefficients to shift the minimum to the specified point. The bilevel optimization can be written as follows:
\begin{problem}{opt:poly}
	\begin{array}{rl}
		\min\limits_{\bm{\theta}} & (x^*(\bm{\theta}) -\bar{x})^2 + (y(x^*(\bm{\theta}),\bm{\theta})-\bar{y})^2\\
		s.t.& x^*(\bm{\theta}) = \mbox{arg}\min\limits_{x} y(x,\bm{\theta}).
	\end{array}
\end{problem}

Though this example does not have direct application in robotics, it is a simplified analogue of robotics problems in reinforcement learning, where an overparameterized value function must be learned to achieve a specific task.

\subsubsection{Problem Parameters}

In this example, we set the target global minimum to $(\bar{x}, \bar{y})=(1.7, 7.3)$ and initialize the polynomial coefficients as shown in Table~\ref{tab:poly_coeff}.

\begin{table}[H]
	\vspace*{\pretablespace}
	\caption{Initial Polynomial Coefficients}\label{tab:poly_coeff}
	\centering
    \begin{tabular}{|c|c|c|c|c|c|c|}
    \hline
    $\theta_0$ & $\theta_1$ & $\theta_2$ & $\theta_3$ & $\theta_4$ & $\theta_5$& $\theta_6$ \\
    \hline
    10.0 & 2.6334 & -4.3443&  0.0  &  0.8055& -0.1334&  0.0389\\
    \hline
    \end{tabular} 
\end{table}

\subsubsection{Inner Optimization}\label{sec:poly-inner}

We consider two different methods for solving the inner optimization. The first method is a standard non-linear gradient descent (GD) method \rev{applied directly to the (unconstrained) polynomial function}. On the first outer iteration, this method requires an initialization point, $x_0$, but subsequent (inner) optimizations are initialized using the previous minimum. Since we are dealing with a polynomial, \rev{the Jacobian} can be computed analytically using the IFT.

The second method is a tight SDP relaxation of the inner optimization, which is implemented using our SDPRLayer.
We parameterize \rev{the} cost matrix as
\renewcommand{\arraystretch}{1.1}
\begin{equation*}
	\bm{Q}_{\bth} = \begin{bmatrix}
		\theta_0 & \frac{1}{2}\theta_1 & \frac{1}{3}\theta_2 & \frac{1}{4}\theta_3 \\ 
        \frac{1}{2}\theta_1 & \frac{1}{3}\theta_2 & \frac{1}{4}\theta_3 & \frac{1}{3}\theta_4 \\ 
        \frac{1}{3}\theta_2 & \frac{1}{4}\theta_3 & \frac{1}{3}\theta_4 & \frac{1}{2}\theta_5 \\ 
        \frac{1}{4}\theta_3 & \frac{1}{3}\theta_4 & \frac{1}{2}\theta_5 & \theta_6 \\ 
    \end{bmatrix}.
\end{equation*}
\renewcommand{\arraystretch}{1}
The following constraints lead to a tight semidefinite relaxation:
\begin{equation*}
	\bm{A}_0 = \begin{bmatrix}
		1 & 0 & 0 & 0 \\
		0 & 0 & 0 & 0 \\
		0 & 0 & 0 & 0 \\
		0 & 0 & 0 & 0 \\
		\end{bmatrix}, \quad \bm{A}_1 = \begin{bmatrix}
			0 & 0 & \frac{1}{2} & 0 \\
			0 & -1 & 0 & 0 \\
			\frac{1}{2} & 0 & 0 & 0 \\
			0 & 0 & 0 & 0 \\
			\end{bmatrix},
\end{equation*}
\begin{equation*}
	\bm{A}_2 = \begin{bmatrix}
		0 & 0 & 0 & 1 \\
		0 & 0 & -1 & 0 \\
		0 & -1 & 0 & 0 \\
		1 & 0 & 0 & 0 \\
		\end{bmatrix}, \quad \bm{A}_3 = \begin{bmatrix}
			0 & 0 & 0 & 0 \\
			0 & 0 &  0 & \frac{1}{2} \\
			0 & 0 & -1 & 0 \\
			0 & \frac{1}{2} & 0 & 0 \\
			\end{bmatrix}.
\end{equation*}

Note that at least one of the constraints is redundant for the original QCQP.\footnote{This can be seen by noting that the original QCQP variable is given by $\bx^{\top}=\begin{bmatrix}1 & x & x^2 & x^3 \end{bmatrix}$, which can be enforced by two constraints and one homogenizing constraint.}

\subsubsection{Outer Optimization}

Since the outer optimization is unconstrained, we find the minimum using gradient descent, with the gradients being computed based on the solution map of the inner optimization. The optimization is terminated when the loss function has a value less than $1\times10^{-4}$.

\subsubsection{Results}

The progression of the bilevel optimization is shown in Figure~\ref{fig:poly_evo_ex}. The key observation is that when gradient descent is used for the inner optimization, the convergence of the overall optimization to a valid solution depends heavily on the initialization point. 

When gradient descent is initialized at $x_0=2$ (left plot), the solution converges to and remains at a local minimum of the polynomial. The gradients that are computed here correspond to the local minimum solution, thus the outer optimization then shifts the \emph{local minimum} of the polynomial to the target point, while the global minimum remains largely unchanged. 

When gradient descent is initialized at $x_0=-2$ (center plot), the solution initially converges to the global minimum and provides valid gradients. As the outer iterations proceed, the inner optimization temporarily gets stuck in a local minimum. At this point, the outer optimization pushes the new global minimum in the wrong direction until the inner optimization eventually converges to the new global minimum. Subsequently, gradient descent is able to reach a valid solution, though it \rev{required an increased number of iterations}.
This case shows that initializing the inner optimization well does not guarantee that it will not get stuck in a local minimum at some point during the optimization, hence providing incorrect gradient information. 

In stark contrast to gradient descent, the SPDRLayer \emph{always} converges to the global solution of the problem and therefore always provides the correct gradients to the outer optimization (right plot). As a consequence, the outer optimization is able to consistently push the global minimum towards the target, and is even able to switch discontinuously to a different minimum as necessary (see Figure~\ref{fig:poly_evo_ex}).

\subsection{Stereo \rev{Localization} Example}\label{sec:stereo_exp}

In this example, \rev{we investigate the performance of SDPRLayers on a stereo-vision localization problem, which commonly needs to be solved when estimating the state of a robot.\footnote{Localization is also sometimes referred to as pointcloud regression.} We assume that we} have stereo camera measurements of a known set of $N_m$ features, \rev{with known} data association and no outliers. Pixel measurements are converted into Euclidean measurements using the following differentiable \rev{inverse-}stereo-camera model (see~\citet[Section \MakeUppercase{\romannumeral 3}C]{gridsethKeepingEyeThings2022} for details):
\begin{equation}\label{eqn:inv_stereo}
	\bm{m}_k = \frac{b}{d_k}
	\begin{bmatrix}
	u_{k} - c_u \\
	\frac{f_u}{f_v}(v_{k} - c_v)\\
	f_u
	\end{bmatrix},
\end{equation}
where $\bm{m}_k\in\mathbb{R}^3$ is the Euclidean measurement of the $k^{th}$ feature, $b$ is the camera baseline, $f_u$ and $f_v$ are the horizontal and vertical focal lengths for the cameras, respectively, and $c_u$ and $c_v$ are the horizontal and vertical centers for the cameras, respectively. The left-camera horizontal, vertical, and disparity measurements (in pixels) of the $k^{th}$ feature are given by $u_{k}$, $v_{k}$, and $d_k$, respectively. 

The (pixel-space) measurements are assumed to be perturbed by Gaussian noise with standard deviations of $\sigma_u$ and $\sigma_v$ in the horizontal and vertical directions, respectively. In turn, the Euclidean measurements are also perturbed by noise,
\begin{equation}
	\tilde{\bm{m}}_k = \bm{m}_k + \bm{\epsilon}_k,
\end{equation}
where $\bm{\epsilon}_k$ is a approximated by a zero-mean, Gaussian noise term with anisotropic covariance matrix $\bm{\Sigma}_{k}$. This matrix is a function of the camera parameters and can be computed using the inverse measurement model, as shown in~\cite{holmesSemidefiniteRelaxationsMatrixWeighted2024}. 
The maximum-likelihood camera pose can be found by solving the matrix-weighted localization problem \cite{holmesSemidefiniteRelaxationsMatrixWeighted2024}:
\begin{problem}{opt:mwex_inner}
	\begin{array}{rl}
		\min\limits_{\bm{C},\bm{t}} &\sum\limits_{k=1}^{N_m} \bm{e}_{k}^{\top} \bm{W}_{k}(\bm{\theta}) \bm{e}_{k} \\
		\mbox{s.t.} &\bm{C}\in\mbox{SO}(3), \\
		& \bm{e}_{k} = \tilde{\bm{m}}_k(\bm{\theta}) - \bm{C}(\bm{m}_k + \bm{t}) ,
	\end{array}
\end{problem}
where $\bm{\theta}$ represents the camera baseline, $\bm{C}$ and $\bm{t}$ are the camera rotation and translation, $\bm{m}_k$ is a feature point with known location, $\tilde{\bm{m}}_k(\bm{\theta})$ is the Euclidean measurement of the point from the camera frame and $\bm{W}_{k}(\bm{\theta})$ is the inverse of the covariance matrix, $\bm{\Sigma}_k$.  

\subsubsection{Problem Parameters}

\rev{The experiments in this section use the setup shown in Figure~\ref{fig:stereo_setup}.}
We ran simulated experiments using the stereo camera parameters given in Table~\ref{tab:cam_params}. The features, $\bm{m}_k$, were arranged in an equally spaced, $8$-by-$8$ grid occupying a 1.0 m by 1.0 m rectangle located at the origin (similar to a checkerboard~\rev{calibration pattern}). In each experiment, \rev{ground-truth} camera poses were placed randomly at a radius of 3 m from the center of the grid, within a 90 degree cone. Orientations of the camera poses were also randomized, but it was ensured that the center of the grid points were within a 90 degree field of view of the camera. 

\begin{table}[H]
	\vspace*{\pretablespace}
	\caption{Ground-Truth Camera Parameters}\label{tab:cam_params}
	\centering
    \begin{tabular}{ |l|c|c|c|c|c|c|c| }
    \hline
    Parameter& $b $ & $f_u$ & $f_v$ & $c_u$ & $c_v$ & $\sigma_u$ &  $\sigma_v$ \\ \hline
	Units& m & $\frac{\mbox{pix}}{\mbox{m}}$ & $\frac{\mbox{pix}}{\mbox{m}}$ & pix & pix & pix & pix \\
	\hline
    Values & 0.24 & 484.5  & 484.5  & 0.0  &  0.0 & 0.5 &  0.5 \\
	\hline
    \end{tabular} 
\end{table}

\subsubsection{Solving the Optimization}

\begin{figure}[t]
    \centering
    \includegraphics[width=\linewidth]{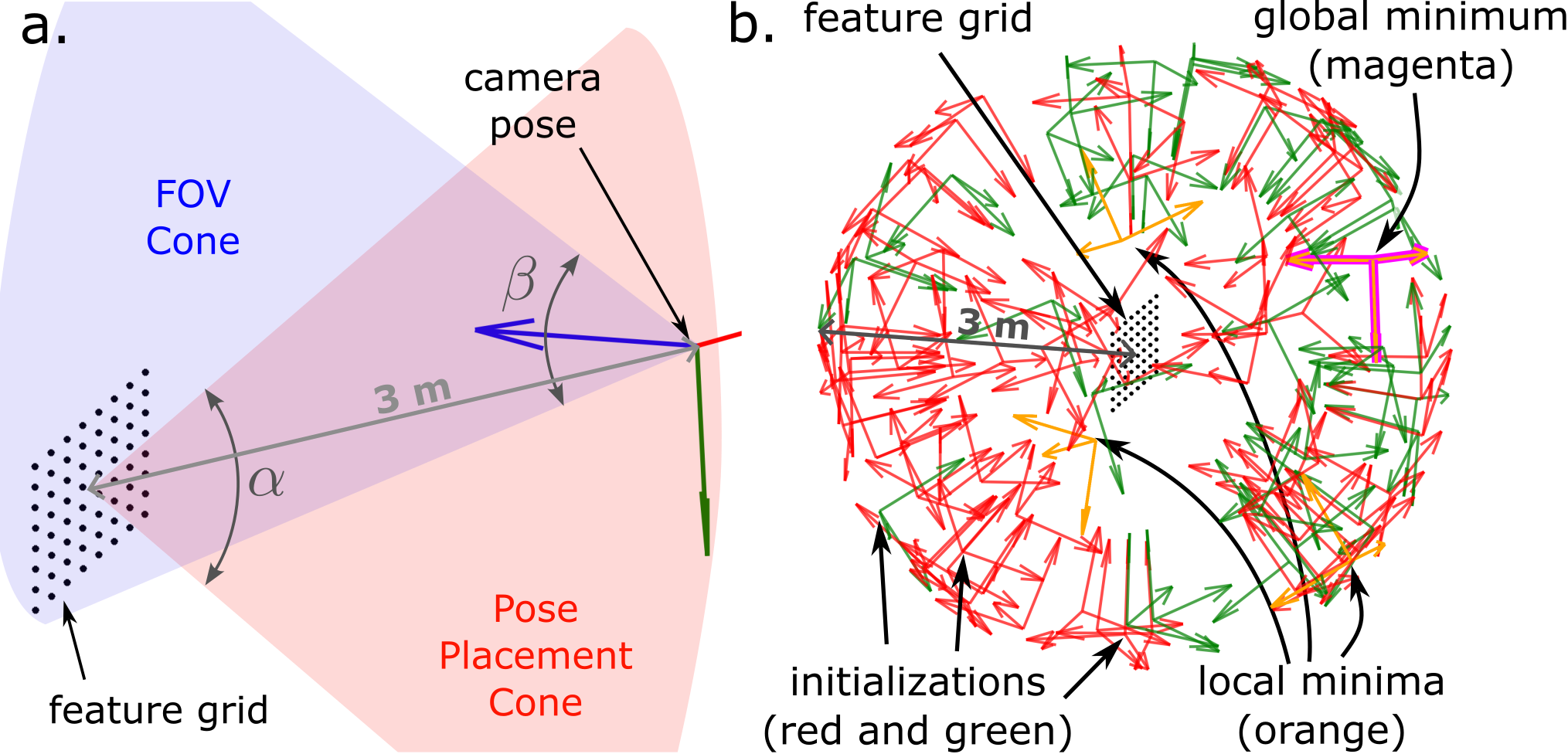}
    \caption{a. Setup for one of the ground-truth poses used in the stereo tuning example. The position of the ground-truth poses are placed 3 m from the center of the grid in a cone of angle $\beta = 90~\mbox{deg}$ (red cone). The poses are oriented such that the center of the grid is within the field of view (FOV) of the camera ($z$) axis, $\alpha=90~\mbox{deg}$. b. 100 random initialization samples for Theseus' Gauss-Newton solver. The initializations are colour coded based on whether they converged to the local minimum (red) or global minimum (green). The local and global solutions are orange and magenta frames, respectively.}
	\label{fig:stereo_setup}
\end{figure}

Since Problem~\eqref{opt:mwex_inner} is a non-linear least-squares problem, with $\mbox{SO}(3)$ Lie group constraints, it can be solved using the Theseus optimization framework~\cite{pineda2022theseus}. Moreover, as shown by~\citet{holmesSemidefiniteRelaxationsMatrixWeighted2024}, Problem~\eqref{opt:mwex_inner} also has a tight semidefinite relaxation, which is particularly robust to noise when redundant constraints are used.  

We implemented both Theseus and the SDPRLayer to solve the optimization. Since Theseus is a local solver, it requires an initial estimate of the pose to solve Problem~\eqref{opt:mwex_inner}. Similar to the polynomial example, after the first iteration, the inner optimization is warm-started with the previous solution. We investigated situations in which Theseus is initialized with both the ground-truth and randomized initializations in order to demonstrate the effect of convergence to local minima. The random initializations used in this context are exemplified in Figure~\ref{fig:stereo_setup}b; translations are randomly selected on the surface of a 3 m ball around the feature grid (at the origin), with orientation selected such that the $z$-axis points at the center of the grid and the other two axes are random. Figure~\ref{fig:stereo_setup}b shows that roughly 60\% the initializations converge to poor local minima. 

In contrast, the SDPRLayer does not require any initialization and always finds the globally optimal solution.

\subsubsection{\rev{Jacobian Comparisons}}\label{sec:jac-compare}

\rev{

\begin{table*}[t]
	\color{black}
	\centering
	\caption{\rev{Jacobian Comparison for Solution ($\bm{C},\bm{t}$) w.r.t. Feature Locations ($\bm{m}_k$) for Scalar-Weighted Localization (Relative To SVD Solution)}}\label{tbl:jac-scl-results}
	\begin{tabular}{lccccc}
		\toprule
		Method & Jac. Diff. (mean) & Jac. Diff. (std) & RMSE Trans. & RMSE Rot. & Backprop Time (s) \\
		\midrule
		SDPR-SDP & 1.88E-05 & 2.44E-05 & 6.09E-07 & 1.08E-08 & \textbf{1.36E-01} \\
		SDPR-CIFT & \textbf{1.29E-06} & \textbf{3.06E-06} & 6.09E-07 & 1.08E-08 & 3.89E-01 \\
		SDPR-IS (Default) & 3.10E-06 & 7.82E-06 & 6.09E-07 & 1.08E-08 & 2.08E-01 \\
		\midrule
		Theseus-GT & 1.02E-02 & 6.62E-03 & 1.61E-15 & 9.80E-10 & 1.86E-01 \\
		Theseus-GT-UR & 8.28E-06 & 1.19E-05 & 4.81E-14 & 4.42E-08 & 7.74E-01 \\
		Theseus-RND & 1.02E-02 & 6.62E-03 & \textbf{1.57E-15} & \textbf{1.05E-09} & 1.80E-01 \\
		\bottomrule
	\end{tabular}
\end{table*}

\begin{table*}[t]
	\color{black}
	\centering
	\caption{\rev{Jacobian Comparison for Solution ($\bm{C},\bm{t}$) w.r.t. Feature Locations ($\bm{m}_k$) for Matrix-Weighted Localization (Relative To Theseus-GT-UR Solution)}}\label{tbl:jac-mw-results}
	\begin{tabular}{lccccc}
		\toprule
		Method & Jac. Diff. (mean) & Jac. Diff. (std) & RMSE Trans. & RMSE Rot. & Backprop Time (s)\\
		\midrule
		SDPR-SDP & 1.49E-04 & 3.78E-04 & 8.38E-06 & 9.53E-06 & \textbf{1.33E-01} \\
		SDPR-CIFT & \textbf{3.11E-05} & \textbf{7.01E-05} & 8.38E-06 & 9.53E-06 & 4.27E-01 \\
		SDPR-IS (Default) & 3.95E-05 & 8.31E-05 & 8.38E-06 & 9.53E-06 & 2.31E-01 \\
		\midrule
		Theseus-GT & 2.07E-02 & 2.88E-02 & \textbf{6.10E-10} & \textbf{7.45E-09} & 1.42E-01 \\
		Theseus-RND & 1.95E-01 & 2.60E-01 & 9.02E-01 & 6.08E-01 & 1.79E-01 \\
		\bottomrule
	\end{tabular}
	\vspace*{5pt}
	\caption*{\rev{(LEGEND) \textbf{SDPR-SDP}: SDP solution diff. through SDP (CVXPYLayers); \textbf{SDPR-CIFT}: SDP solution diff. through QCQP using the classic IFT (Section~\ref{sec:backprop-cift}); \textbf{SDPR-IS}: SDP solution diff. through QCQP using implicit selections (Section~\ref{sec:backprop-sift}); \textbf{Theseus-GT}: Theseus solution diff. implicitly with ground-truth init.; \textbf{Theseus-RND}: Theseus solution diff. implicitly with random init.; \textbf{Theseus-GT-UR}: Theseus solution diff. via unrolling with ground-truth init.}}
\end{table*}

To demonstrate the validity of our approach we first compare the Jacobians of the solution to Problem~\eqref{opt:mwex_inner} with respect to the parameters for different differentiation approaches across 50 trials. To make the experiments in this section as accurate as possible, we used Mosek~\cite{apsMOSEKOptimizerAPI2024} to solve the SDP with tolerances set to $1\times10^{-12}$ and set the Theseus tolerances to $1\times10^{-12}$. 

We performed an initial experiment with scalar weighting ($\bm{W}_k=\bm{I}$), since this allows the problem to be solved using the (differentiable) Singular Value Decomposition (SVD). The SVD solution is both closed-form and differentiable, serving as an accurate baseline for comparison of our approaches (see \citet{umeyamaLeastSquaresEstimationTransformation1991} for the details). We compute the relative difference between Jacobians using the infinity norm:
\begin{equation}
	\Delta \bm{J}_{\rm est} = \frac{\lVert \bm{J}_{\rm est} - \bm{J}_{\rm true} \rVert_{\infty}}{\lVert \bm{J}_{\rm true} \rVert_{\infty}},
\end{equation}
where $\bm{J}_{\rm true}$ is the SVD Jacobian and $\bm{J}_{\rm est}$ is the Jacobian of the alternate method.
In Table~\ref{tbl:jac-scl-results} we show the mean (Jac. Diff. (mean)) and standard deviation (Jac. Diff. (std)) of $\Delta \bm{J}_{\rm est}$ across trials for each approach. We also provide the root mean squared error (RMSE) of the relative translation and rotation between the solutions to demonstrate that the all solutions converge to (approximately) the same point. \footnoterev{Relative translation and rotation was computed in the Lie algebra vector space.}

The three SDPR approaches have (approximately) the same deviation in the Jacobian, indicating that they are equivalent in terms of accuracy. Despite having better accuracy in terms of the actual solution, the Jacobians of the local methods were less accurate except when using the `unrolling' method (UR).\footnoterev{Unrolling refers to backpropagation of gradients through all iterations of an optimization.}

When the matrix weights were introduced, the SVD solution is no longer applicable, and we instead perform our comparisons with respect to the unrolled Theseus solution initialized at the ground truth (Theseus-GT-UR), since it was the most accurate. The results for this comparison are shown in Table~\ref{tbl:jac-mw-results}. We call attention to the fact that, when Theseus is initialized randomly (Theseus-RND) in the matrix-weighted case, it can converge to local minima (as shown in Figure~\ref{fig:stereo_setup}b) and, as a result, the Jacobian does not match well with the ideal case. On the other hand, the SDPR solutions always converge to global minima and have accurate Jacobians.

Tables~\ref{tbl:jac-scl-results} and~\ref{tbl:jac-mw-results} also include the mean backpropagation time for each case. As discussed in Section~\ref{sec:theory}, reusing the multipliers and certificate matrix when backpropagating (SDPR-IS) leads to faster compute times than recomputing them (SDPR-CIFT). Curiously, backpropagation through the SDP KKT conditions (SDPR-SDP) was consistently faster than differentiating through the QCQP conditions. We posit that this is due to the small size of the SDP in this example and the fact that the CVXPYLayers backend is implemented using optimized C++ libraries while our QCQP differentiation is implemented using standard Python libraries.

}
\subsubsection{\rev{Baseline Calibration}}

\rev{We now provide a further example of how gradient information obtained from an uncertified local solution can contaminate the processes that rely on gradient information. We again consider a bilevel optimization that uses Problem~\eqref{opt:mwex_inner} to }calibrate the stereo baseline\rev{, $b$,} of a camera rig. 
Problem~\eqref{opt:mwex_inner} constitutes the inner optimization, \rev{while the} loss minimized by the outer optimization is the squared error between estimated camera pose and ground-truth camera pose,
\begin{problem}{opt:mwex_outer}
	\begin{array}{rl}
		\min\limits_{\bm{\theta}} & \Vert \bm{t}^*(\bm{\theta}) - \bm{t}_{gt} \Vert_2^2 + \Vert \bm{C}^*(\bm{\theta})^{\top} \bm{C}_{gt} - \bm{I} \Vert_F^2
	\end{array},
\end{problem}
where  $ \left\{\bm{C}^*(\bm{\theta}),\bm{t}^*(\bm{\theta})\right\} $ is the estimated pose from the inner optimization and $ \left\{\bm{C}_{gt},\bm{t}_{gt}\right\} $ represents the ground-truth pose. In practice, the loss could be unsupervised (i.e., not include ground-truth data), but we use this simplified loss to make comparison between approaches more straightforward. Since the outer optimization is unconstrained it can be solved iteratively using stochastic gradient descent.

\begin{figure}[t]
    \centering
    \includegraphics[width=\linewidth]{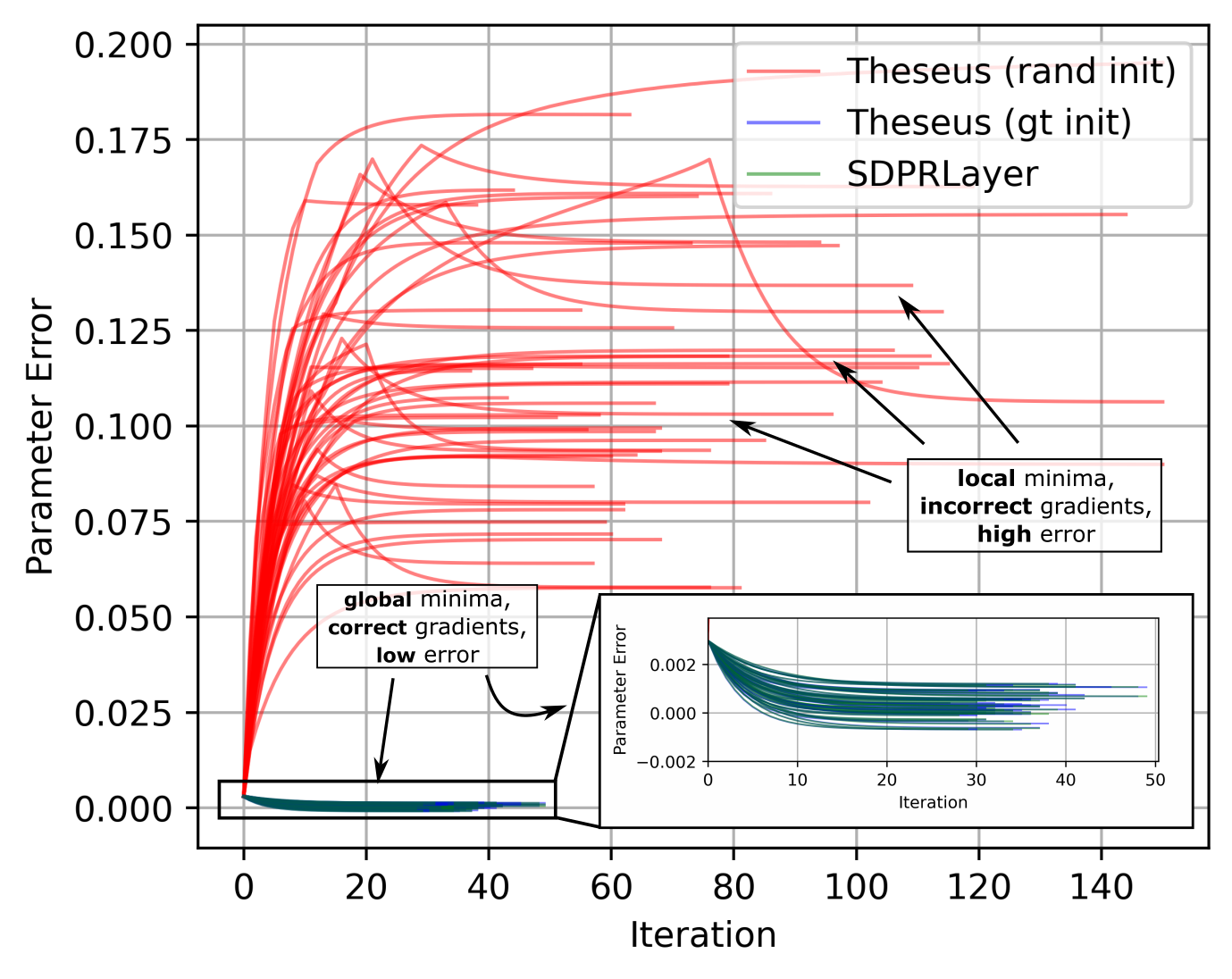}
    \caption{Baseline parameter error trajectories across outer optimization iterations for 50 trials with different inner loop optimization approaches. Theseus with ground-truth initialization and SDPRLayer always converge to global minima, hence provide \emph{correct} gradients to the outer opt. Theseus with random initialization sometimes converges to local minima and therefore provides \emph{incorrect} gradients to the outer optimization, resulting in large parameter error.}
	\label{fig:param_traj}
\end{figure}

We ran 50 stereo calibration experiments with \rev{20} different ground-truth poses and compared the results between Theseus and our approach. In the Theseus implementation, we use a Gauss-Newton solver with a stepsize of $0.2$, with termination tolerances set to $1\times10^{-8}$, and \rev{implicit differentiation for backpropagation. For this experiment, the backend of the SDPRLayer implementation used the SCS to solve the SDP with tolerance set to $1\times10^{-9}$~\cite{odonoghueConicOptimizationOperator2016}.}

The baseline parameter was initialized with a 0.003 m error from the true value and the outer optimization was solved using stochastic gradient descent with a learning rate of $1\times10^{-4}$. Each experiment terminated when the outer optimization gradient magnitude was less than $1\times10^{-3}$ or 150 iterations had been reached. We stress that the \emph{only difference} in the trials using Theseus in this section is the initialization used.

The parameter error trajectories for the different approaches are shown in Figure~\ref{fig:param_traj} across outer optimization iterations and aggregate results are provided in Table~\ref{tbl:baseline_results}. From Figure~\ref{fig:param_traj}, it is clear that initializing randomly causes convergence to camera baseline values that are completely incorrect. By contrast, both our approach and Theseus with ground-truth initialization converge to a low level of error in a shorter number of iterations. Note that the individual trajectories of these two approaches are very close to each other. This is confirmed by the investigation of the gradients and inner optimization error. This is to be expected, since both approaches converge to the global minimum and thus return almost exactly the same gradients (subject to numerical precision of solvers). 


\begin{table}[t]
	\caption{Baseline Tuning Results Across Runs}\label{tbl:baseline_results}
	\color{black}
	\begin{tabularx}{0.49\textwidth}{lXXXXXX}
		\toprule
		Method & \raggedright Final Baseline Error (avg) & \raggedright Final Baseline Error (std) & \raggedright Number of Outer Iterations (avg) &  Outer Loss (avg) \\
		\midrule
		Theseus-RND & 1.132e-01 & 3.191e-02 & 8.034e+01 & 1.384e+02 \\
		Theseus-GT & \textbf{3.463e-04} & \textbf{4.901e-04} & \textbf{3.540e+01} & \textbf{9.854e-03} \\
        SDPR-IS & \textbf{3.454e-04} & \textbf{4.907e-04} & \textbf{3.422e+01} & \textbf{9.857e-03} \\
        \bottomrule
	\end{tabularx}
\end{table}

\begin{figure*}[t]
    \centering
    \includegraphics[width=\textwidth]{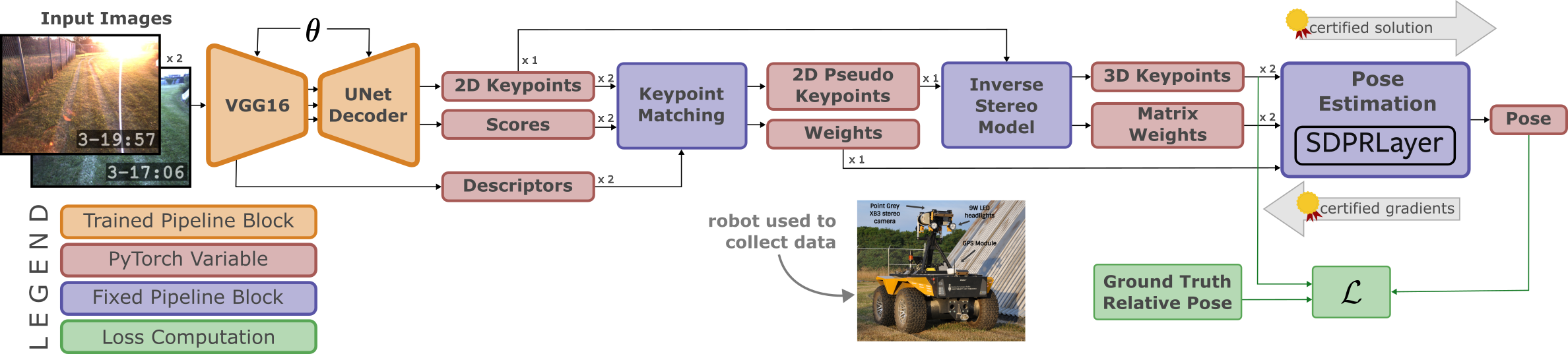}
    \caption{Diagram of our Pytorch pipeline (based on the pipeline in \cite{gridsethKeepingEyeThings2022}) that we use to estimate the relative pose between two images (i.e., localize a target image to a (stored) source image). Orange blocks denote the deep neural networks whose weights are tuned during training (VGG16 network and UNet Decoder network). Blue blocks denote blocks that are part of the pipeline, but do not have trainable parameters. Red blocks indicate key PyTorch tensor variables. Green blocks and arrows indicate training loss function computation.}
	\label{fig:pipeline}
\end{figure*}

Our investigations suggest that the divergence of the randomly initialized Theseus approach is exactly because the inner optimization converges to local minima for some poses. In turn, this leads to gradients that push the baseline in the wrong direction. Initial experimentation suggests that the `kinks' in the red trajectories of Figure~\ref{fig:param_traj} occur because some poses of the batch are able to escape local minima and converge to better solutions as the baseline parameter changes.

The aggregate results in Table~\ref{tbl:baseline_results} corroborate our findings in Figure~\ref{fig:param_traj}; random initialization leads to poor tuning overall, increasing the number of required iterations, the average error and variation of tuned parameter. Again, our approach and Theseus with ground-truth initialization match very closely.

\subsection{Deep Learned Visual Features for Localization}\label{sec:deeplearn}

Our goal in this section is to demonstrate that the SDPRLayer can be used to train deep neural networks for large, real-world robotics problems. We consider the task of supervised learning of visual features for robot localization when environment lighting conditions can change drastically. In particular, we adopt and modify the learning pipeline introduced by \citet{gridsethKeepingEyeThings2022} for stereo-vision-based robot localization and show that the SDPRLayer enables \emph{matrix-weighted} localization, which leads to increased accuracy. 

Our adaptation of this fully differentiable pipeline is shown in Figure \ref{fig:pipeline}. A convolutional neural network (VGG16 and UNet Decoder) is used to extract keypoints along with descriptors and scores for source and target RGB, stereo-images (i.e., four images total). 2D keypoints are detected in the left source and target images and matched. The keypoints in each stereo-image pair are then converted to 3D coordinates using a stereo camera model and disparity between left and right images. Finally, the 3D matched coordinates are used to compute a relative pose between source and target with a differentiable pose estimator. Relative (scalar) weighting of the keypoint pairs in the estimation are determined based on the scores and descriptor alignment (see (5) in \cite{gridsethKeepingEyeThings2022}).

The pose-estimation stage of the original pipeline used a Singular Value Decomposition (SVD) since\rev{, as mentioned above, }it provides a closed-form and differentiable solution. However, since this method only supports scalar weights in the cost function, it cannot fully incorporate the (directional) uncertainty of keypoints that are derived from stereo images~\cite{matthiesErrorModelingStereo1987}.

Properly accounting for depth uncertainty has been shown to be important for accurate state estimation in robotics \cite{matthiesErrorModelingStereo1987,wengMotionStructureEstimation1992,maimoneTwoYearsVisual2007}. 
This can be accomplished by replacing the pose-estimation block of the baseline pipeline with the (non-convex) matrix-weighted pose estimation given in Problem~\eqref{opt:mwex_inner}. As shown in Section~\ref{sec:stereo_exp}, solving this optimization with local solvers can subject the network training to erroneous gradient information. To avoid this issue, we solve the optimization using the SDPRLayer with Mosek as the internal solver. As before, the matrix weights are computed based on the inverse covariance of each 3D keypoint, which is known based on camera model and assumed pixel-space covariance.\footnote{We assumed an isotropic, pixel-space, noise distribution with a standard deviation of 0.5 pixels.} The matrix weights are also scaled by the scalar weights provided by the matching block in the pipeline. 

Finally, similar to \citet{chenSelfSupervisedFeatureLearning2023}, we replace the encoder segment of the neural network with a VGG16 network \cite{simonyanVeryDeepConvolutional2015} (truncated at \texttt{conv\_5\_3} layer) that has been pretrained on ImageNet \cite{dengImageNetLargescaleHierarchical2009} to facilitate faster training.  We also retrained the baseline (SVD) pose estimation with the VGG16 encoder network to ensure a fair comparison.

\subsubsection{Training}

We kept as many parameters unchanged as possible between the baseline training setup and our setup. The feature detector network was trained for 30 epochs (10000 samples in each epoch) on an NVIDIA Tesla V100 DGXS GPU using the same keypoint and pose-estimation training loss (with the same relative weights) as in \cite{gridsethKeepingEyeThings2022} (see (7) and (8) therein). The pose-estimation loss was not used for a `warm-up' period of 10 epochs (i.e., only keypoint loss in this period).

For training, we used 100000 training samples and 20000 validation samples from the `In-The-Dark' dataset\footnote{Dataset publically available at \url{http://asrl.utias.utoronto.ca/datasets/2020-vtr-dataset/}.}, which was also used for training by \citet{gridsethKeepingEyeThings2022} and has many instances of severe lighting changes. Using the Adam optimizer with a learning rate of $1\times10^{-4}$, the decoder network was trained from scratch and the VGG16 network was fine-tuned from its pretrained weights.

\subsubsection{Testing}

It is assumed that ground-truth data is unavailable during inference. As such, a random sample consensus (RANSAC) was used in the matching block to find the correct set of inliers based on reprojection error. In all cases, the pose estimation within the RANSAC algorithm was performed using SVD to minimize runtime, with the final pose refinement performed with either SVD (scalar weighted) or SDPR (matrix weighted).

We compare our pipeline (`Ours') against the baseline pipeline (`Baseline') proposed in \cite{gridsethKeepingEyeThings2022} and the baseline with VGG16 (`Baseline (w. VGG16)') on a set of test runs that were held out of the training and validation data. In particular, we tested the localization pipelines on all source-target stereo-image pairs\footnote{The image data association between different runs was also performed using the code from \cite{gridsethKeepingEyeThings2022}.} from all combinations of runs 2, 11, 16, 17, 23, 28, and 35 from the In-The-Dark dataset and assessed the error based on comparison with ground truth. Test runs for the baseline were performed using the available code and network weight parameters associated with \cite{gridsethKeepingEyeThings2022}.\footnote{Code publically available at \url{https://github.com/utiasASRL/deep_learned_visual_features}} 

The original test set involved only small relative-pose changes between source and target viewpoints since the localization pose graph was quite dense. As such, the original ground-truth relative-pose changes were 6.36 cm and 0.56 deg on average in translation and rotation, respectively. However, we also generated a more challenging test set by forcing the relative-pose change between map and live frames to be strictly greater than either 0.5 m translationally or 4 degrees rotationally. For this challenging dataset, the average relative-pose change was 55.6 cm and 1.39 deg in translation and rotation, respectively.

\begin{figure*}[t]
    \centering
    \includegraphics[width=\textwidth]{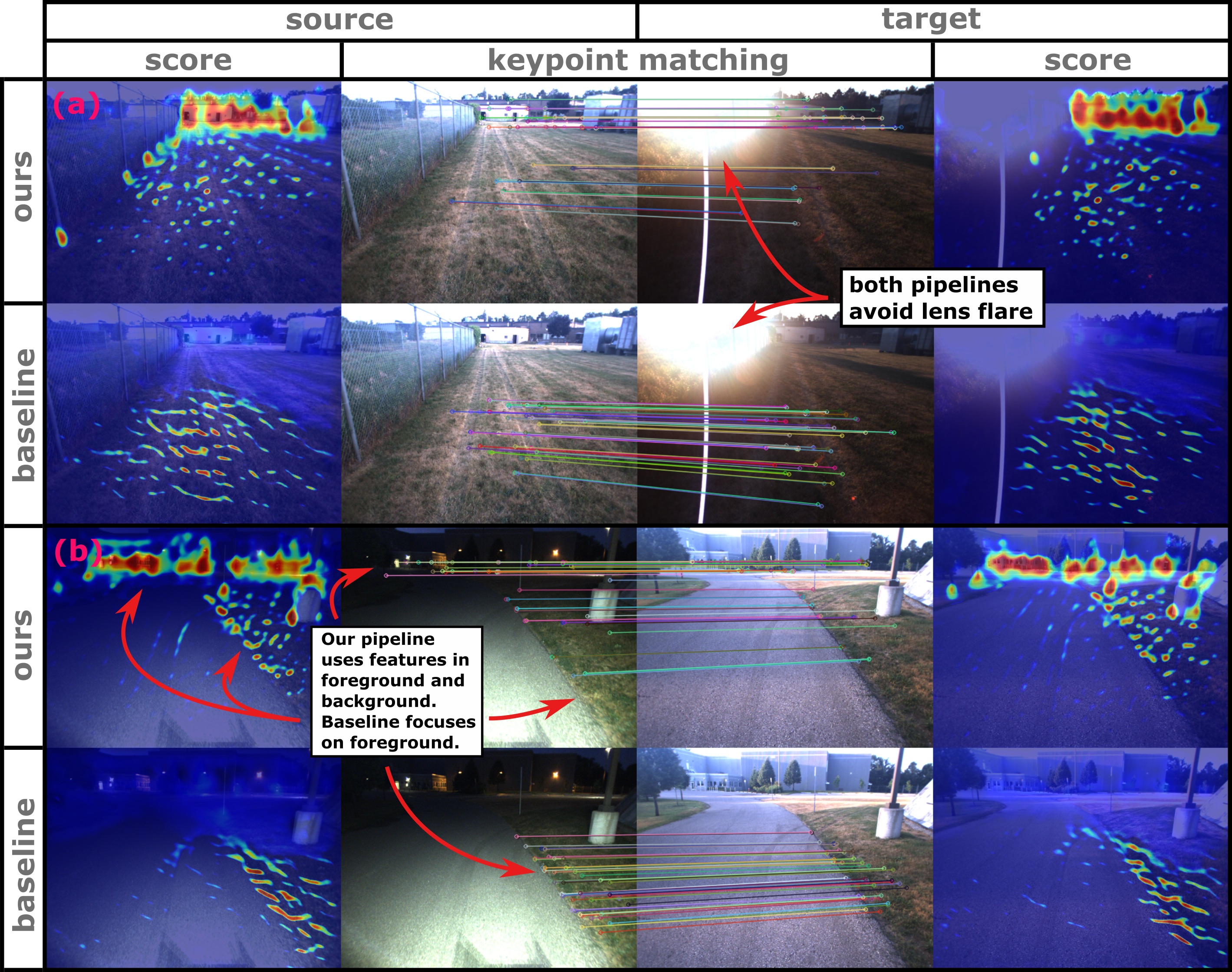}
    \caption{Scores and keypoint matches for two selected stereo-image pairs from the In-The-Dark dataset after RANSAC filtering. Image pair (a) corresponds to source (Run 27, Frame 2182) and target (Run 8, Frame 1886) and demonstrates `lens flare'. Image pair (b) corresponds to source (Run 21, Frame 2057) and target (Run 27, Frame 1830) and demonstrates a night-day match. The keypoint-match images (center two columns) display the source and target images with the 50 highest-weighted keypoint matches overlaid. The score columns show the score map from the neural network corresponding to their adjacent images.}
	\label{fig:feature_match}
\end{figure*}

\subsubsection{Results}

Tables \ref{tab:results_summary} and~\ref{tab:results_summary_challenge} show number of inliers (from RANSAC), longitudinal position error, lateral position error, heading (yaw) error,\footnote{All errors are average RMSE values relative to ground-truth vehicle pose frame from \cite{patonBridgingAppearanceGap2016a}.} and inference time per pose, averaged across all runs. In both test sets, it is clear that our pipeline performs best in terms of \rev{longitudinal, lateral, and heading error, although the difference in heading error is negligible}. We posit that the improved longitudinal and lateral error are due to appropriate accounting of the stereo-based depth uncertainty when matrix weights are used. 

The average inference time differs between the two tables because the challenging dataset typically requires more RANSAC iterations to converge to an acceptable solution. The baseline pipeline with VGG16 took substantially longer than with the original encoder from~\cite{gridsethKeepingEyeThings2022}, likely due to larger network size of VGG16. Interestingly, we note that, since the SDP used to globally solve Problem~\eqref{opt:mwex_inner} is relatively small, the average inference time with the SDPRLayer was only \rev{about two times} longer than inference with the closed-form, SVD-based pose estimation.

\begin{table}
	\caption{Summary of Average Inliers and Localization Error Across Test Data from \cite{gridsethKeepingEyeThings2022}}
	\label{tab:results_summary}
	\centering
	\color{black}
	\begin{tblr}{lrrrrr}
		\hline
		{Pipeline} &  {Avg.\\ Inliers} & {Long.\\ Pos.\\ Err. (m)} &  { Lat. \\Pos.\\ Err. (m) } & {  Head. \\Err. \\(deg)} & {Avg. Inf.\\ Time\\ (s)} \\
		\hline
		{Baseline}      &          \textbf{547.6} & 0.209 & 0.134 & 0.414 &    \textbf{0.052}  \\
		{Baseline\\(w. VGG16)} & 541.6 & 0.125 & 0.049 & 0.301 &    0.133  \\
		Ours     & 535.6 & \textbf{0.035} & \textbf{0.021} & \textbf{0.288} & 0.243 \\
		\hline
	\end{tblr}
\end{table}

\begin{table}
	\caption{Summary of Average Inliers and Localization Error Across Test Runs from Challenging Dataset (Trans. $\geq 0.5 $ m or Rot $\geq 4$ deg)}
	\label{tab:results_summary_challenge}
	\centering
	\color{black}
	\begin{tblr}{lrrrrr}
		\hline
		{Pipeline} &  {Avg.\\ Inliers} & {Long.\\ Pos.\\ Err. (m)} &  { Lat. \\Pos.\\ Err. (m) } & {  Head. \\Err. \\(deg)} & {Avg. Inf.\\ Time\\ (s)} \\
		\hline
		{Baseline}      &    \textbf{446.3} & 0.432 & 0.144 & 0.377 &  \textbf{0.089}\\
		{Baseline\\(w. VGG16)}      &   431.3 & 0.213 & 0.072 & 0.318 &    0.182 \\
		Ours & 423.4 & \textbf{0.053} & \textbf{0.027} & \textbf{0.316} & 0.327 \\
		\hline
	\end{tblr}
\end{table}

The qualitative performance of the pipelines (with VGG16 encoder) is assessed in Figure \ref{fig:feature_match}, which provides the scores and most-highly weighted keypoint matches on two samples from the In-the-Dark dataset. Both pipelines successfully find a good set of keypoint matches and manage to avoid problems induced by drastic lighting changes (i.e., `lens flare' and night-day matching).

We note that the baseline consistently focuses on keypoints in the foreground. We speculate that this is because, during training, the network cannot separate the high depth uncertainty from the low lateral uncertainty in faraway measurements and therefore reduces the weighting on the background keypoints. 

On the other hand, our pipeline characterizes this directional uncertainty using the matrix weights. It is therefore able to weight the background keypoints more heavily than the baseline. We note in passing that this can be quite advantageous in localization because keypoints that are farther away typically provide better information about orientation and, as long as the problem is matrix weighted, will not corrupt the translation estimate. Both pipelines seem to use the background keypoints to achieve good heading estimates, but our implementation does so without adversely affecting the translation estimates.

The experiments shown in this section clearly demonstrate that our SDPRLayer can be used to train neural networks in real-world robotics pipelines. The bottleneck of the SDPRLayer approach is computational cost of solving the SDP. However, this localization problem can be solved quickly due to its small size and is useful in a practical context since it can provide globally optimal image registration with a large of numbers of features.

\section{Conclusion and Future Work}\label{sec:conclusion}

We have presented the SDPRLayer, a differentiable optimization layer for polynomial optimization problems with tight semidefinite relaxations. We have demonstrated that differentiable optimization approaches with \emph{local} solvers \rev{can provide gradient information that does not correspond to the global solution (due to convergence to spurious local minima)}. By extension, the training/optimization process may be lengthened or even fail entirely to achieve its objective.  On the other hand, we provided theoretical and experimental results showing that the SDPRLayer \rev{efficiently computes} \emph{certified} gradients, in the sense that they correspond to the certified, global minimum of the optimization problem. 

The first two examples shown in this paper have demonstrated the potential pitfalls of naive application of differentiable optimization. The final example demonstrates that the SDPRLayer can be used for real-world robotics applications that combine deep-learned and model-based components. This method could be readily extended to train more recent feature-detection-and-matching pipelines (e.g., SuperGlue~\cite{sarlinSuperGlueLearningFeature2020b} or its variants~\cite{lindenbergerLightGlueLocalFeature2023,jiangOmniGlueGeneralizableFeature2024}) with pose registration error.

\rev{Currently, our theory is limited to SDP relaxations that are exactly tight (i.e., solution has a rank of one). For problems that do not have such relaxations, we have suggested an alternative approach using backpropagation through the KKT conditions of the relaxation (Section~\ref{sec:recourse}), but note that this approach is subject to theoretical limitations. An interesting direction of future work includes alleviating these limitations as well as further experimentation with problems that are not exactly tight.}

\rev{Another current limitation of our approach is that the forward (optimization) and backward steps currently} take place on the CPU, leading to costly memory transfers when \rev{combined with} training with a GPU. 
\rev{A parallel GPU implementation of our approach} necessary \rev{for} the application of SDPRLayers -- and more generally, certifiable methods -- to larger robotics problems.

Finally, we believe that SDPRLayers could be applied to other areas of robotics apart from the perception methods demonstrated in this paper. In particular, differentiable Model Predictive Control is an exciting problem that has been recently studied in the literature \cite{romeroActorCriticModelPredictive2023} and may benefit from certifiable gradients in practice.


\appendix

\rev{
\subsection{Jacobians of KKT Conditions}\label{sec:kkt-jac-params}

In this section, we derive the expressions for the Jacobian of the KKT conditions with respect to the (vectorized) input parameters $\{\bm{Q}_{\bth}, \bm{A}_{\bth i}\}$. Throughout this section, we adopt the notion of \emph{differentials} and notation from~\citet{magnusMatrixDifferentialCalculus2019}. We will make use of the following properties:
\begin{proposition}\label{thm:diffvec-prop}
	Let $\bm{B}\in \mathbb{R}^{n\times n}$ and let $\bm{a},\bm{b}\in\mathbb{R}^n$ be fixed vectors. Then we have the following differential relationships:
	\begin{gather}
		\dd(\bx^{\top}\bm{B}\bx) = \vect{\bx\bx^{\top}}^{\top} \vect{\dd\bm{B}} \label{eqn:app-prop1}\\
		\dd(\bm{B}\bx) = (\bm{I}\otimes\bx^{\top}) \vect{\dd\bm{B}}\label{eqn:app-prop2}
	\end{gather}
\end{proposition}
\begin{IEEEproof}
	The first property can be verified easily using the trace operator and its relation to vectorized matrices, $\trace{\bm{A}^{\top}\bm{B}} = \vect{\bm{A}}^{\top}\vect{\bm{B}}$. We have,
	\begin{equation*}
		\begin{split}
		\dd(\bx^{\top}\bm{B}\bx) &= \trace{\bx^{\top}\dd\bm{B}\bx} = \trace{\bx\bx^{\top}\dd\bm{B}} \\&= \vect{\bx\bx^{\top}}^{\top} \vect{\dd\bm{B}}
		\end{split}
	\end{equation*}
	We consider the second property element-wise:
	\begin{equation*}
		\left(\dd(\bm{B}\bx)\right)_i = \dd(\bm{e}_i\bm{B}\bx) = \vect{\bm{e}_i\bx^{\top}} \vect{\dd\bm{B}},
	\end{equation*}
	where $\bm{e}_i$ is a vector of zeros with a one at index $i$. Collecting the differentials into a vector, we have:
	\begin{equation*}
		\dd(\bm{B}\bx) = \begin{bmatrix}
			\vect{\bm{e}_1\bx^{\top}}\\\vdots\\\vect{\bm{e}_n\bx^{\top}}
		\end{bmatrix}\vect{\dd\bm{B}} = (\bm{I}\otimes\bx^{\top}) \vect{\dd\bm{B}}
	\end{equation*}
\end{IEEEproof}

We note that the differential of the certificate matrix is given by:
\begin{equation}\label{eqn:cert-mat-diff}
	\dd\bm{H} = \dd\bm{Q}_{\bth}+ \sum\limits_{i=1}^{\ncon}\dd\bm{A}_{\bth i}\lambda_i
\end{equation}
Applying Proposition~\ref{thm:diffvec-prop}, the differential of the KKT conditions in~\ref{eqn:kkt-conditions} are given by
\begin{gather*}
	\dd(2\bm{H}\bx) = 2(\bm{I}\otimes\bx^{\top})(\vect{\dd\bm{Q}_{\bth}} + \sum\limits_{i=1}^{\ncon}\lambda_i\vect{\dd\bm{A}_{\bth i}}),\\
	\dd(\bx^{\top}\bm{A}_{\bth i}\bx) = \vect{\bx\bx^{\top}}\vect{\dd\bm{A}_{\bth i}}.
\end{gather*}
Letting 
\begin{equation}
	\dd\bm{\nu}^{\top} = \begin{bmatrix}\vect{\dd\bm{Q}_{\bth}}^{\top} & \vect{\dd\bm{A}_{\bth 1}}^{\top} & \cdots & \vect{\dd\bm{A}_{\bth m}}^{\top} \end{bmatrix},	
\end{equation}
The differential of the KKT conditions is given by
\begin{equation}
	\dd\bm{k}(\bm{z},\bth) = \begin{bmatrix}
		\dd(2\bm{H}\bx) \\ 
		\dd(\bx^{\top}\bm{A}_{\bth 1}\bx)\\
		\vdots\\
		\dd(\bx^{\top}\bm{A}_{\bth m}\bx)\\
		0
	\end{bmatrix}.
\end{equation}
We have the following differential relationship (when $\bm{z}$ is constant),
\begin{equation*}
	\dd\bm{k}(\bm{z},\bth) = \bm{N} \dd\bm{\nu},
\end{equation*}
where, applying~\eqref{eqn:app-prop1},~\eqref{eqn:app-prop1}, and~\eqref{eqn:cert-mat-diff}, the Jacobian is given by
\begin{equation}
	\bm{N} = \begin{bmatrix}
		2(\bm{I}\otimes\bx^{\top}) & \blam'^{\top} \otimes(2\bm{I}\otimes\bx^{\top}) \\ 
		\bm{0} & \bm{I} \otimes\vect{\bx\bx^{\top}} \\
		\bm{0} & \bm{0}
	\end{bmatrix},
\end{equation}
where $\blam'$ is the vector of Lagrange multipliers with $\lambda_0$ removed.
}
\subsection{Addressing Tightness}\label{sec:address_tightness}

For the convenience of the user, we have added two functions to the SDPRLayer module to address the tightness of the SDP solution. The first function (\texttt{check\_tightness}) computes the ratio of the maximum two eigenvalues of the SDP solution,
\begin{equation*}
	r = \frac{\lambda_{1}(\bm{X}^*(\bth))}{\lambda_{2}(\bm{X}^*(\bth))},
\end{equation*}
where, for $\bm{A}\in\mathbb{R}^{\nvar}$, $\lambda_{i}(\bm{A})$ denotes the $i^{th}$ eigenvalue of $\bm{A}$, where $\lambda_{1}\geq \dots \geq \lambda_{\ncon}$.
If the ratio exceeds a given threshold then the solution can be considered to be rank-1, and the relaxation is tight. Empirically, we have found a ratio of $r =1\times10^\rev{5}$ to be a good indicator of relaxation tightness.

If the specified SDP relaxation is not initially tight, the method proposed by \citet{dumbgenGloballyOptimalState2024} can be used to find a set of redundant constraints that may tighten the problem. We introduce a second function (\texttt{find\_constraints}) that can be used to find all possible constraints for a particular problem instance, using the \emph{AutoTight} approach from~\cite{dumbgenGloballyOptimalState2024}. If the relaxation is still not tight after adding these constraints, then the \emph{only}\footnote{By \emph{completeness} of AutoTight, see Section IV-E of~\cite{dumbgenGloballyOptimalState2024}.} way to tighten the relaxation is to modify the formulation (i.e., cost or variable selection).

The following procedure can be used to find tight relaxations to POPs:

\begin{enumerate}
	\item A given POP of some variable, $\bm{z}\in\mathbb{R}^{\nvar}$, can always be reformulated as Problem~\eqref{opt:QCQPh} (or, alternatively, Problem~\eqref{opt:QCQPh}) using the approach outlined in \citet{carloneEstimationContractsOutlierRobust2023} or \citet{yangCertifiablyOptimalOutlierRobust2023}. Roughly speaking, this involves selecting a set of monomials of $\bm{z}$ that are capable of representing the problem in QCQP form. These monomials are collected in a new variable $\bx$, and constraints are added to enforce the relationship between $\bx$ and $\bm{z}$.
	\item Once formulated as a QCQP, the relaxed Problem~\eqref{opt:SDPR} can be solved and the rank of the solution can be tested. The solution and the rank test can be performed using our SDPRLayers module. \label{step:check_tight}
	\item If the relaxation is not tight, then the iterative procedure outlined by \citet{dumbgenGloballyOptimalState2024} can be used to find a tight relaxation:
	\begin{enumerate}
		\item \rev{Find} all possible redundant constraints for the QCQP. These constraints can be gradually added until the relaxation becomes tight. We provide a function to do this with the SDPRLayer.
		\item If all constraints have been added and the relaxation is still not tight, then additional \emph{variables} can be added to $\bx$ by increasing the degree of the monomials used in the formulation.\footnote{This step is akin to ascending the \rev{Moment-SOS hierarchy} \cite{yangCertifiablyOptimalOutlierRobust2023}.} The procedure then continues from Step \ref{step:check_tight}.
	\end{enumerate}
\end{enumerate}

\begin{figure}[!t]
    \centering
    \includegraphics[width=0.8\linewidth]{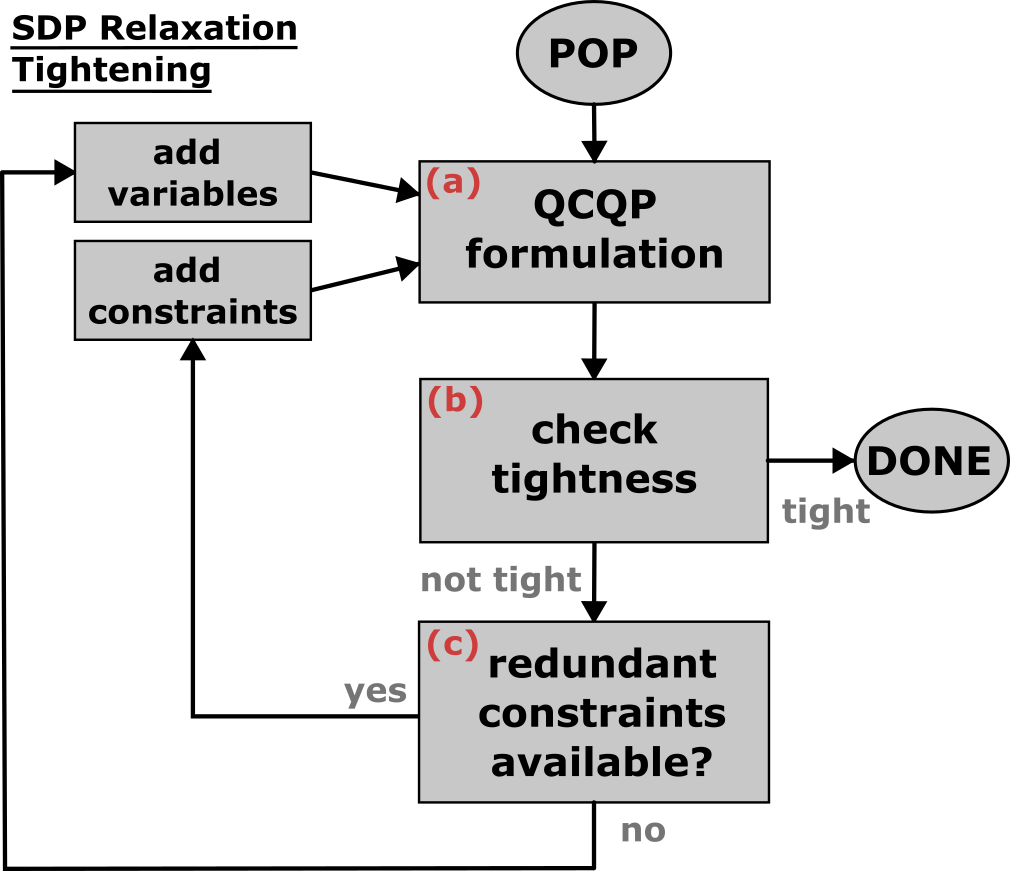}
    \caption{General algorithm for tightening an SDP relaxation of a POP. Given a POP, (a) a QCQP can be formulated using the methods discussed in~\cite{yangCertifiablyOptimalOutlierRobust2023} or~\cite{carloneEstimationContractsOutlierRobust2023}. (b) Tightness of the relaxation can be assessed using the SDPRLayer presented herein (Section \ref{sec:address_tightness}). (c) If the relaxation is not tight, SDPRLayer provides a method to find all possible constraints to improve tightness based on \cite{dumbgenGloballyOptimalState2024}. If the relaxation is still not tight, then variables can be added (i.e., move to a higher level of Lasserre's hierarchy).}
	\label{fig:tighen_algo}
\end{figure}

Figure \ref{fig:tighen_algo} provides a visual representation of this procedure. \rev{Note that the approach outlined here is similar to the Moment-SOS hierarchy. However, the hierarchy does not necessarily find all possible constraints at each a given level before proceeding to the next level and adding new variables. 

As mentioned,} there are some guarantees on when this procedure -- or more generally, \rev{the Moment-SOS hierarchy} -- results in a tight relaxation, but it can also lead to intractably large SDPs. Finding efficient, tight relaxations for POPs remains an active area of research.

\subsection{Modifications to CVXPYLayers}\label{sec:cvxpy-mods}
CVXPYLayers parses user-provided DCPs into a canonical form and solves the DCP using a convex program solver (default solver is Splitting Conic Solver or SCS~\cite{odonoghueConicOptimizationOperator2016}). This canonicalization can often lead to inefficiencies when converting a DCP to a cone program. We have observed this empirically for the SDPs studied in robotics, depending on the solver to be used. This issue can be avoided by formulating the problem in dual form, but the implementation of CVXPYLayers did not previously support differentiation of slack or dual variables. As such, we have modified the interface of CVXPYLayers (and its underlying dependencies) to expose the (differentiable) Lagrange multipliers and slack variables as outputs, which are already computed by the underlying solvers. 

We also extended the interface to allow the requisite primal, dual, and slack solution variables to be provided directly by the user. In this case, the forward pass simply injects the solution variables from the external solver, bypassing the optimization. On the backward pass, the stored variables are used by the existing implicit differentiation machinery. This allows users to opt to use \emph{external solvers} rather than the solvers in the CVXPYLayers codebase. We have used this approach to solve and differentiate SDPs with Mosek \cite{apsMOSEKOptimizerAPI2024}, which is often faster and more robust than the default solver.

\bibliographystyle{plainnat}
\bibliography{LearningViaConvexRelaxations}

\end{document}